\documentclass[runningheads]{llncs}

\usepackage[width=122mm,left=12mm,paperwidth=146mm,
  height=193mm,top=12mm,paperheight=217mm]{geometry}
\usepackage{graphicx,amsmath,amssymb,subcaption,cite,floatrow}
\usepackage[colorlinks=true,bookmarks=false]{hyperref}
\newcommand{\tss}[1]{\textsuperscript{#1}}
\newcommand{\app}{\raise.17ex\hbox{$\scriptstyle\sim$}}

\newlength\savewidth\newcommand\shline{\noalign{\global\savewidth\arrayrulewidth
  \global\arrayrulewidth 1pt}\hline\noalign{\global\arrayrulewidth\savewidth}}

\begin{document}
\pagestyle{headings}\mainmatter
\title{Learning to Refine Object Segments}
\titlerunning{Learning to Refine Object Segments}
\def\ECCV16SubNumber{755}
\newcommand{\affiliations}{\scriptsize Authors contributed equally to this work while at FAIR. Current affiliations: Pedro~O.~Pinheiro is with the Idiap Research Institute and Ecole Polytechnique F\'ed\'erale de Lausanne (EPFL); Tsung-Yi~Lin is with Cornell University and Cornell Tech.}
\author{Pedro O. Pinheiro\thanks{\affiliations}, Tsung-Yi Lin$^\star$, Ronan Collobert, Piotr Doll\'ar}
\authorrunning{Pinheiro, Lin, Collobert, Doll\'ar}
\institute{Facebook AI Research (FAIR)}
\maketitle

%%%%%%%%%%%%%%%%%%%%%%%%%%%%%%%%%%%%%%%%%%%%%%%%%%%%%%%%%%%%%%%%%%%%%%%%%%%%%%%%%%%%%%%%%%%%%%%%%%%
\begin{abstract}
Object segmentation requires both object-level information and low-level pixel data. This presents a challenge for feedforward networks: lower layers in convolutional nets capture rich spatial information, while upper layers encode object-level knowledge but are invariant to factors such as pose and appearance. In this work we propose to augment feedforward nets for object segmentation with a novel top-down refinement approach. The resulting bottom-up/top-down architecture is capable of efficiently generating high-fidelity object masks. Similarly to skip connections, our approach leverages features at all layers of the net. Unlike skip connections, our approach does not attempt to output independent predictions at each layer. Instead, we first output a coarse `mask encoding' in a feedforward pass, then refine this mask encoding in a top-down pass utilizing features at successively lower layers. The approach is simple, fast, and effective. Building on the recent DeepMask network for generating object proposals, we show accuracy improvements of 10-20\% in average recall for various setups. Additionally, by optimizing the overall network architecture, our approach, which we call SharpMask, is 50\% faster than the original DeepMask network (under .8s per image).
\end{abstract}

%%%%%%%%%%%%%%%%%%%%%%%%%%%%%%%%%%%%%%%%%%%%%%%%%%%%%%%%%%%%%%%%%%%%%%%%%%%%%%%%%%%%%%%%%%%%%%%%%%%
\section{Introduction}\label{sec:introduction}

As object detection~\cite{Felzenszwalb2010, SermanetICLR2013, Szegedy15, He2014sppNet, Girshick2014rcnn, girshick15fastrcnn, RenNIPS15fasterRCNN, bell15ion} has rapidly progressed, there has been a renewed interest in object instance segmentation~\cite{mscoco2015}. As the name implies, the goal is to both detect and segment each individual object. The task is related to both object detection with bounding boxes~\cite{mscoco2015, Everingham10, imagenet_cvpr09} and semantic segmentation~\cite{ShottonJC08, Everingham10, FarabetPAMI2013, PinheiroICML2014, eigen2015predicting, zheng2015conditional, ChenICLR15, schwing2015fully, noh2015learning}. It involves challenges from both domains, requiring accurate pixel-level object segmentation coupled with identification of each individual object instance.

\begin{figure}[t]\centering
  \includegraphics[width=.99\textwidth]{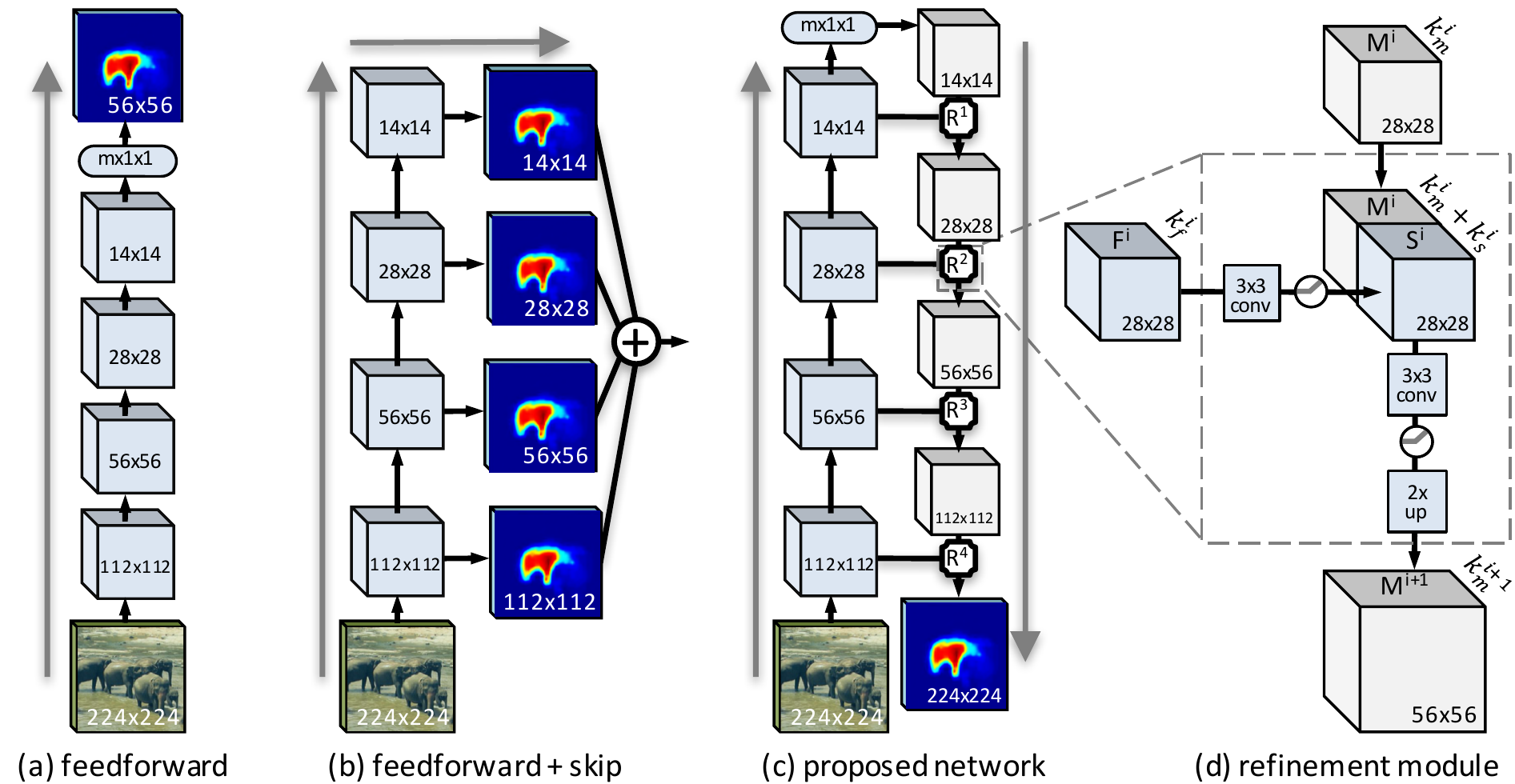}
\caption{Architectures for object instance segmentation. (a) Feedforward nets, such as DeepMask~\cite{pinheiro2015learning}, predict masks using only upper-layer CNN features, resulting in coarse pixel masks. (b) Common `skip' architectures are equivalent to making independent predictions from each layer and averaging the results~\cite{longCVPR15fcn, BharathCVPR2015, Xie2015hed}, such an approach is not well suited for object instance segmentation. (c,d) In this work we propose to augment feedforward nets with a novel top-down refinement approach. The resulting bottom-up/top-down architecture is capable of efficiently generating high-fidelity object masks.}
\label{fig:refinement}
\end{figure}

A number of recent papers have explored the use convolutional neural networks (CNNs) \cite{lecun1998gradient} for object instance segmentation~\cite{hariharan14sds, pinheiro2015learning, he2015segmentation, BharathCVPR2015}. Standard feedforward CNNs~\cite{AlexNet, Simonyan15, GoogLeNet, he2015deep} interleave convolutional layers (with pointwise nonlinearities) and pooling layers. Pooling controls model capacity and increases receptive field size, resulting in a coarse, highly-semantic feature representation. While effective and necessary for extracting object-level information, this general architecture results in low resolution features that are invariant to pixel-level variations. This is beneficial for classification and identifying object instances but poses challenge for pixel-labeling tasks. Hence, CNNs that utilize only upper network layers for object instance segmentation~\cite{hariharan14sds, pinheiro2015learning, he2015segmentation}, as in Figure~\ref{fig:refinement}a, can effectively generate coarse object masks but have difficulty generating pixel-accurate segmentations.

%##################################################################################################
%##################################################################################################

For pixel-labeling tasks such as semantic segmentation and edge detection, `skip' connections~\cite{sermanet-cvpr-2013, longCVPR15fcn, BharathCVPR2015, Xie2015hed}, as shown in Figure~\ref{fig:refinement}b, are popular. In practice, common skip architectures are equivalent to making independent predictions from each network layer and upsampling and averaging the results (see Fig.~2 in~\cite{BharathCVPR2015}, Fig.~3 in~\cite{longCVPR15fcn}, and Fig.~3 in~\cite{Xie2015hed}). This is effective for semantic segmentation as local receptive fields in early layers can provide sufficient data for pixel labeling. For object segmentation, however, it is necessary to differentiate between object instances, for which local receptive fields are insufficient (e.g. local patches of sheep fur can be labeled as such but without object-level information it can be difficult to determine if they belong to the same animal).

In this paper, we propose a novel CNN which efficiently merges the spatially rich information from low-level features with the high-level object knowledge encoded in upper network layers. Rather than generating independent outputs from multiple network layers, our approach first generates a coarse \emph{mask encoding} in a feedforward manner, which is simply a semantically meaningful feature map with multiple channels, then refines it by successively integrating information from earlier layers. Specifically, we introduce a \emph{refinement module} and stack successive such modules together into a top-down refinement process. See Figures~\ref{fig:refinement}c and \ref{fig:refinement}d. Each refinement module is responsible for `inverting' the effect of pooling by taking a mask encoding generated in the top-down pass, along with the matching features from the bottom-up pass, and merging the information in both to generate a new mask encoding with double the spatial resolution. The process continues until full resolution is restored and the final output encodes the object mask. The refinement module is efficient and fully backpropable.

We apply our approach in the context of object proposal generation~\cite{AlexePAMI12, Uijlings13, ZitnickD14, ARXIV2015MCG, KrahenbuhlK14, humayun2014, Hosang2015proposals}. The seminal object detection work on R-CNN~\cite{Girshick2014rcnn} follows a two-phase approach: first, an object proposal algorithm is used to find regions in images that may contain objects; second, a CNN assigns each proposal a category label. While originally object proposals were constructed from low-level grouping and saliency cues~\cite{Hosang2015proposals}, recently CNNs have been adopted for this task~\cite{Szegedy15, RenNIPS15fasterRCNN, pinheiro2015learning}, leading to massive improvements in detection accuracy. In particular, Pinheiro et al.~\cite{pinheiro2015learning} demonstrated how to adopt a CNN to generate rich object instance segmentations in an image. The proposed model, called DeepMask, predicts how likely an image patch is to fully contain a centered object and also outputs an associated segmentation mask for the object, if present. The model is run convolutionally to generate a dense set of object proposals for an image. DeepMask outperforms previous object segment proposal methods by a substantial margin~\cite{pinheiro2015learning}.

In this work we utilize the DeepMask architecture as our starting point for object instance segmentation due to its simplicity and effectiveness. We augment the basic DeepMask architecture with our refinement module (see Figure~\ref{fig:refinement}) and refer to the resulting approach as \emph{SharpMask} to emphasize its ability to produce sharper, higher-fidelity object segmentation masks. In addition to the top-down refinement, we also revisit the basic bottom-up architecture of the DeepMask network and likewise optimize it for the segmentation task.

SharpMask improves segmentation mask quality relative to DeepMask. For object proposal generation, average recall on the COCO dataset~\cite{mscoco2015} improves 10-20\% and establishes the new state-of-the-art on this task. Moreover, we optimize our core architecture and improve speed by 50\% with respect to DeepMask, with an average of .76s per image. Our fast model, which still outperforms previous results, runs at .46s, or, by using additional image scales, we can boost small object recall by \app$2\times$. Finally we show SharpMask proposals substantially improve object detection results when coupled with the Fast R-CNN detector~\cite{girshick15fastrcnn}.

The paper is organized as follows: \S\ref{sec:related} presents related work, \S\ref{sec:refinement} introduces our novel top-down refinement network, \S\ref{sec:architecture} describes optimizations to the network architecture, and finally \S\ref{sec:exp} validates our approach experimentally.

All source code for reproducing the methods in this paper will be released.

%%%%%%%%%%%%%%%%%%%%%%%%%%%%%%%%%%%%%%%%%%%%%%%%%%%%%%%%%%%%%%%%%%%%%%%%%%%%%%%%%%%%%%%%%%%%%%%%%%%
\section{Related Work} \label{sec:related}

Following their success in image classification~\cite{AlexNet, Simonyan15, GoogLeNet, he2015deep}, CNNs have been adopted with great effect to pixel-labeling tasks such as depth estimation~\cite{eigen2015predicting}, optical flow~\cite{DosovitskiyICCV2015}, and semantic segmentation~\cite{FarabetPAMI2013}. Below we describe architectural innovations for such tasks, and discuss how they relate to our approach. Aside from skip connections~\cite{sermanet-cvpr-2013, BharathCVPR2015, longCVPR15fcn, Xie2015hed}, which were discussed in \S\ref{sec:introduction}, these techniques can be roughly classified as multiscale architectures, deconvolutional networks, and graphical model networks. We discuss each in turn next. We emphasize, however, that most of these approaches are not applicable to our domain due to severe computational constraints: we must refine hundreds of proposals per image implying the marginal time per proposal must be minimal.

\textbf{Multiscale architectures:} \cite{FarabetPAMI2013, eigen2015predicting, PinheiroICML2014} compute features over multiple rescaled versions of an image. Features can be computed independently at each scale~\cite{FarabetPAMI2013}, or the output from one scale can be used as additional input to the next finer scale~\cite{eigen2015predicting, PinheiroICML2014}. Our approach relies on similar intuition but does not require recomputing features at each image scale. This allows us to apply refinement efficiently to hundreds of locations per image as necessary for object proposal generation.

\textbf{Deconvolutional networks:} \cite{Zeiler10deconv} proposed to invert the pooling process in a CNN to generate progressively higher resolution input images by storing the `switch' variables from the pooling operation. Deconv networks have recently been applied successfully to semantic segmentation~\cite{noh2015learning}. Deconv layers share similarities with our refinement module, however, `switches' are communicated instead of the feature values, which limits the information that can be transferred. Finally, \cite{DosovitskiyICCV2015} proposed to progressively increase the resolution of an optical flow map. This can be seen as a special case of our refinement approach where: (1) the `features' for refinement are set to be the flow field itself, (2) no feature transform is applied to the bottom-up features, and (3) the approach is applied monolithically to the entire image. Restricting our method in any of these ways would cause it to fail in our setting as discussed in \S\ref{sec:exp}.

\textbf{Graphical model networks}: a number of recent papers have proposed integrating graphical models into CNNs by demonstrating they can be formulated as recurrent nets~\cite{zheng2015conditional, ChenICLR15, schwing2015fully}. Good results were demonstrated on semantic segmentation. While too slow to apply to multiple proposals per image, these approaches likewise attempt to sharpen a coarse segmentation mask.

%%%%%%%%%%%%%%%%%%%%%%%%%%%%%%%%%%%%%%%%%%%%%%%%%%%%%%%%%%%%%%%%%%%%%%%%%%%%%%%%%%%%%%%%%%%%%%%%%%%
\section{Learning Mask Refinement}\label{sec:refinement}

We apply our proposed bottom-up/top-down refinement architecture to object instance segmentation. Specifically, we focus on object proposal generation~\cite{Hosang2015proposals}, which forms the cornerstone of modern object detection~\cite{Girshick2014rcnn}. We note that although we test the proposed refinement architecture on the task of object segmentation, it could potentially be applied to other pixel-labeling tasks.

Object proposal algorithms aim to find diverse regions in an image which are likely to contain objects; both  proposal recall and quality correlate strongly with detector performance~\cite{Hosang2015proposals}. We adopt the DeepMask network~\cite{pinheiro2015learning} as the starting point for proposal generation. DeepMask is trained to jointly generate a class-agnostic object mask and an associated `objectness' score for each input image patch. At inference time, the model is run convolutionally to generate a dense set of scored segmentation proposals. We refer readers to~\cite{pinheiro2015learning} for full details.

A simplified diagram of the segmentation branch of DeepMask is illustrated in Figure~\ref{fig:refinement}a. The network is trained to infer the mask for the object located in the center of the input patch. It contains a series of convolutional layers interleaved with pooling stages that reduce the spatial dimensions of the feature maps, followed by a fully connected layer to generate the object mask. Hence, each pixel prediction is based on a complete view of the object, however, its input feature resolution is low due to the multiple pooling stages.

As a result, DeepMask generates masks that are accurate on the object level but only coarsely align with object boundaries, see Figure~\ref{fig:outDeepMask}.  In order to obtain higher-quality masks, we augment the basic DeepMask architecture with our refinement approach. We refer to the resulting method as \emph{SharpMask} to emphasize its ability to produce sharper, pixel-accurate object masks, see Figure~\ref{fig:outSharpMask}. We begin with a high-level overview of our approach followed by further details.

%##################################################################################################
\newcommand{\incp}[2]{\includegraphics[width=.495\textwidth]{figures/#1/#2.jpg}}
\newcommand{\incs}{\hspace{.4mm}}
\begin{figure}[t]
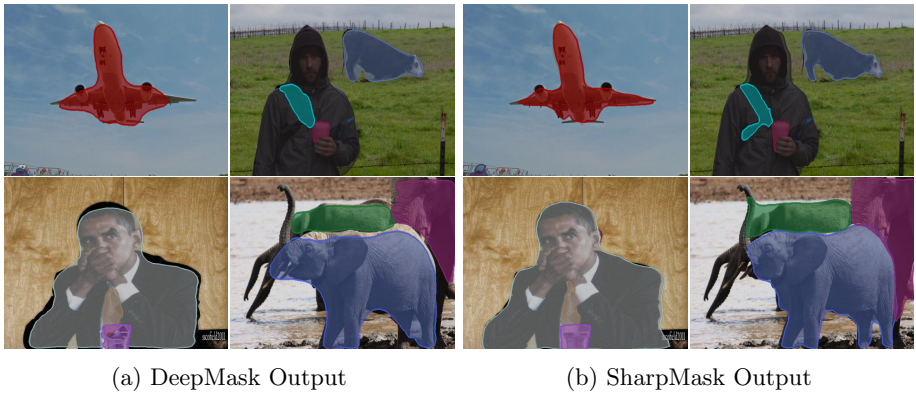
\centering
  \begin{subfigure}[b]{0.495\textwidth}\centering
    \incp{DeepMask}{57429}\incs\incp{DeepMask}{50054}\\
    \incp{DeepMask}{1089}\incs\incp{DeepMask}{757}
    \caption{DeepMask Output}\label{fig:outDeepMask}
  \end{subfigure}
  \begin{subfigure}[b]{0.495\textwidth}\centering
    \incp{SharpMask}{57429}\incs\incp{SharpMask}{50054}\\
    \incp{SharpMask}{1089}\incs\incp{SharpMask}{757}
    \caption{SharpMask Output}\label{fig:outSharpMask}
  \end{subfigure}%
\caption{Qualitative comparison of DeepMask versus SharpMask segmentations. Proposals with highest IoU to the ground truth are shown for each method. Both DeepMask and SharpMask generate object masks that capture the general shape of the objects. However, SharpMask improves the masks near object boundaries.}
\label{fig:comparison}
\end{figure}
%##################################################################################################

\subsection{Refinement Overview}

Our goal is to efficiently merge the spatially rich information from low-level features with the high-level semantic information encoded in upper network layers. Three principles guide our approach: (1) object-level information is often necessary to segment an object, (2) given object-level information, segmentation should proceed in a top-down fashion, successively integrating information from earlier layers, and (3) the approach should invert the loss of resolution from pooling (with the final output matching the resolution of the input).

To satisfy these principles, we augment standard feedforward nets with a top-down refinement process. An overview of our approach is shown in Figure~\ref{fig:refinement}c. We introduce a `refinement module' $R$ that is responsible for inverting the effect of pooling and doubling the resolution of the input mask encoding. Each module $R^i$ takes as input a mask encoding $M^i$ generated in the top-down pass, along with matching features $F^i$ generated in the bottom-up pass, and learns to merge the information to generate a new upsampled object encoding $M^{i+1}$. In other words: $M^{i+1} = R^i(M^i,F^i)$, see Figure~\ref{fig:refinement}d. Multiple such modules are stacked (one module per pooling layer). The final output of our network is a pixel labeling of the same resolution as the input image. We present full details next.

\subsection{Refinement Details}

The feedforward pathway of our network outputs a `mask encoding' $M^1$, or simply, a low-resolution but semantically meaningful feature map with $k_m^1$ channels. $M^1$ serves as the input to the top-down refinement module, which is responsible for progressively increasing the mask encoding's resolution. Note that using $k_m^1>1$ allows the mask encoding to capture more information than a simple segmentation mask, which proves to be key for obtaining good accuracy.

Each refinement module $R^i$ aggregates information from a coarse mask encoding $M^i$ and features $F^i$ from the corresponding layer of the bottom-up computation (we always use the last convolutional layer prior to pooling). By construction, $M^i$ and $F^i$ have the same spatial dimensions; the goal of $R^i$ is to generate a new mask encoding $M^{i+1}$ with double spatial resolution based on inputs $M^i$ and $F^i$. We denote this via $M^{i+1} = R^i(M^i,F^i)$. This process is applied iteratively $n$ times (where $n$ is the number of pooling stages) until the feature map has the same dimensions as the input image patch. Each module $R^i$ has separate parameters, allowing the network to learn stage-specific refinements.

The refinement module aims to enhance the mask encoding $M^i$ using features $F^i$. As $M^i$ and $F^i$ have the same spatial dimensions, one option is to first simply concatenate $M^i$ and $F^i$. However, directly concatenating $F^i$ with $M^i$ poses two challenges. Let $k_m^i$ and $k_f^i$ be the number of channels in $M^i$ and $F^i$ respectively. Typically, $k_f^i$ can be quite large in modern CNNs, so using $F^i$ directly would be computationally expensive. Second, typically $k_f^i \gg k_m^i$, so directly concatenating the features maps risks drowning out the signal in $M^i$.

Instead, we opt to first reduce the number of channels $k_f^i$ (but preserving the spatial dimensions) of these features through a $3\times3$ convolutional module (plus ReLU), generating `skip' features $S^i$, with $k^i_s \ll k^i_f$ channels. This substantially reduces computational requirements, moreover, it allows the network to transform $F^i$ into a form $S^i$ more suitable for use in refinement. An important but subtle point is that during full image inference, as with the features $F^i$, skip features are shared by overlapping image patches, making them highly efficient to compute. In contrast, the remaining computations of $R^i$ are patch dependent as they depend on the local mask $M^i$ and hence cannot be shared across locations.

The refinement module concatenates mask encoding $M^i$ with skip features $S^i$ resulting in a feature map with $k^i_m + k^i_s$ channels, and applies another $3\times3$ convolution (plus ReLU) to the result. Finally, the output is upsampled using bilinear upsampling by a factor of $2$, resulting in a new mask encoding $M^{i+1}$ with $k^{i+1}_m$ channels ($k^{i+1}_m$ is determined by the number of $3\times3$ kernels used for the convolution). As with the convolution for generating the skip features, this transformation is used to simultaneously learn a nonlinear mask encoding from the concatenated features and to control the capacity of the model. Please see Figure~\ref{fig:refinement}d for a complete overview of the refinement module $R$. Further optimizations to $R$ are possible, for details see Figure~\ref{fig:equivalence}.

Note that the refinement module uses only convolution, ReLU, bilinear upsampling, and concatenation, hence it is fully backpropable and highly efficient. In \S\ref{sec:exp:sharp}, we analyze different architecture choices for the refinement module in terms of performance and speed. As a general design principle, we aim to keep $k^i_s$ and $k^i_m$ large enough to capture rich information but small enough to keep computation low. In particular, we can start with a fairly large number of channels but as spatial resolution is increased the number of channels should decrease. This reverses the typical design of feedforward networks where spatial resolution decreases while the number of channels increases with increasing depth.

\subsection{Training and Inference}

We train SharpMask with an identical data definition and loss function as the original DeepMask model. Each training sample is a triplet containing an input patch, a label specifying if the input patch contains a centered object at the correct scale, and for positive samples a binary object mask. The network trunk parameters are initialized with a network that was pre-trained on ImageNet~\cite{imagenet_cvpr09}. All the other layers are initialized randomly from a uniform distribution.

Training proceeds in two stages: first, the model is trained to jointly infer a coarse pixel-wise segmentation mask and an object score, second, the feedforward path is `frozen' and the refinement modules trained. The first training stage is identical to~\cite{pinheiro2015learning}. Once learning of the first stage converges, the final mask prediction layer of the feedforward network is removed and replaced with a linear layer that generates a mask encoding $M^1$ in place of the actual mask output. We then add the refinement modules to the network and train using standard stochastic gradient descent, backpropagating the error only on the horizontal and vertical convolution layers on each of the $n$ refinement modules.

This two-stage training procedure was selected for three reasons. First, we found it led to faster convergence. Second, at inference time, a \emph{single} network trained in this manner can be used to generate either a coarse mask using the forward path only or a sharp mask using our bottom-up/top-down approach. Third, we found the gains of fine-tuning through the entire network to be minimal once the forward branch had converged.

During full-image inference, similarly to~\cite{pinheiro2015learning}, most computation for neighboring windows is shared through use of convolution, including for skip layers $S^i$. However, as discussed, the refinement modules receive a unique input $M^1$ at each spatial location, hence, computation proceeds independently at each location for this stage. Rather than refine every proposal, we simply refine only the most promising locations. Specifically, we select the top $N$ scoring proposal windows and apply the refinement in a batch mode to these top $N$ locations.

To further clarify all implementation details, full source code will be released.

%%%%%%%%%%%%%%%%%%%%%%%%%%%%%%%%%%%%%%%%%%%%%%%%%%%%%%%%%%%%%%%%%%%%%%%%%%%%%%%%%%%%%%%%%%%%%%%%%%%
\section{Feedforward Architecture}\label{sec:architecture}

While the focus of our work is on top-down mask refinement, to obtain a better understanding of object segmentation we also explore factors that effect a feedforward network's ability to generate accurate object masks. In the next two subsections we carefully examine the design of the network `trunk' and `head'.

\subsection{Trunk Architecture}\label{sec:trunk}

We begin by identifying model bottlenecks. DeepMask spends 40\% of its time for feature extraction, 40\% for mask prediction, and 20\% for score prediction. Given the time of feature extraction, increasing model depth or breadth can incur a non-trivial computational cost. Simply upgrading the 11-layer VGG-A model~\cite{Simonyan15} used in \cite{pinheiro2015learning} to the 16-layer VGG-D model can double run time. Recently He et al.~\cite{he2015deep} introduced Residual Networks (ResNet) and showed excellent results. In this work, we use the 50-layer ResNet model pre-trained on ImageNet, which achieves the accuracy of VGG-D but with the inference time of VGG-A. 

We explore models with varying input size \textbf{W}, number of pooling layers \textbf{P}, stride density \textbf{S}, model depth \textbf{D}, and final number of features channels \textbf{F}. These factors are intertwined but we can achieve significant insight by a targeted study.

\textbf{Input size W}: Given a minimum object size O, the input image needs to be upsampled by W/O to detect small objects. Hence, reducing W improves speed of both mask prediction and inference for small objects. However, smaller W reduces the input resolution which in turn lowers the accuracy of mask prediction. Moreover, reducing W decreases stride density S which further harms accuracy.

\textbf{Pooling layers P}: Assuming $2\times2$ pooling, the final kernel width is W/2\tss{P}. During inference, this necessitates convolving with a large W/2\tss{P} kernel in order to aggregate information (e.g., $14\times14$ for DeepMask). However, while more pooling P results in faster computation, it also results in loss of feature resolution.

\textbf{Stride density S}: We define the stride density to be S=W/stride (where typically stride is 2\tss{P}). The smaller the stride, the denser the overlap with ground truth locations. We found that the stride density is key for mask prediction. Doubling the stride while keeping W constant greatly reduces performance as the model must be more spatially invariant relative to a fixed object size.

\textbf{Depth D}: For typical networks~\cite{AlexNet, Simonyan15, GoogLeNet, he2015deep}, spatial resolution decreases with increasing D while the number of features channels F increases. In the context of instance segmentation, reducing spatial resolution hurts performance. One possible direction is to start with lower layers that have less pooling and increase the depth of the model without reducing spatial resolution or increasing F. This would require training networks from scratch which we leave to future work.

\textbf{Feature channels F}: The high dimensional features at the top layer introduce a bottleneck for feature aggregation. An efficient approach is to first apply dimensionality reduction before feature aggregation. We adopt $1\times1$ convolution to reduce F and show that we can achieve large speedups in this manner.

In \S\ref{sec:exp:opt} and Table~\ref{tab:trunk} we examine various choices for W, P, S, D, and F.

\subsection{Head Architecture}\label{sec:head}

%##################################################################################################
\begin{figure}[t]\centering
  \includegraphics[width=.85\textwidth]{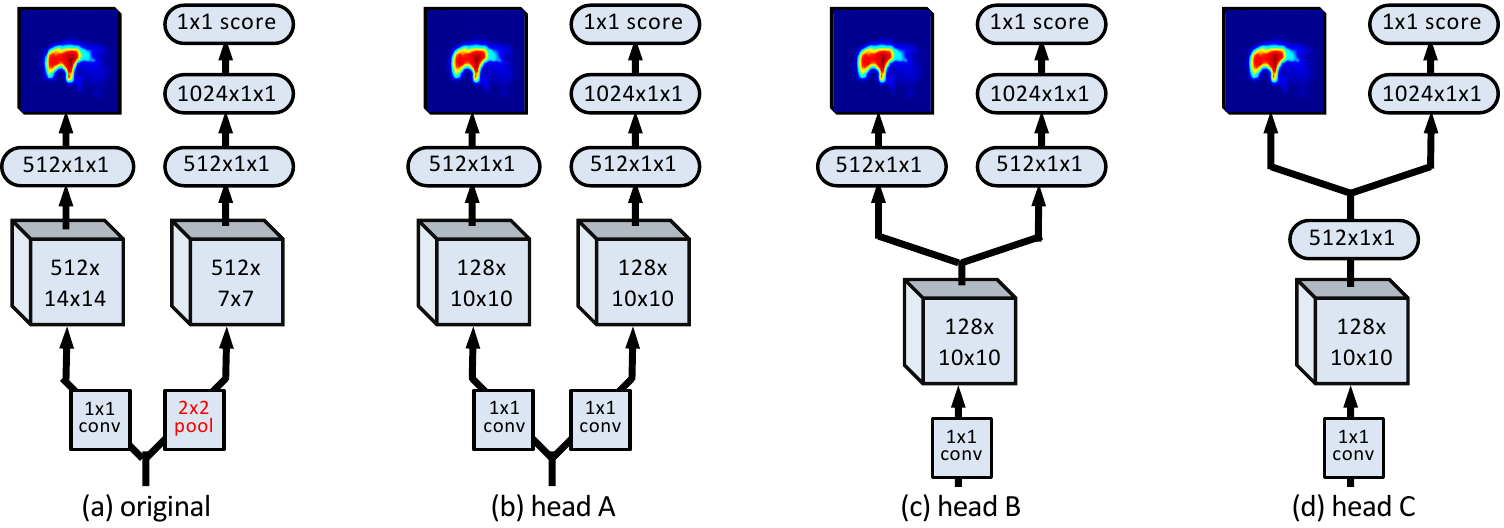}
\caption{Network head architecture. (a) The original DeepMask head. (b-d) Various head options with increasing simplicity and speed. The heads share identical pathways for mask prediction but have progressively simplified score branches.}
\label{fig:head}
\end{figure}
%##################################################################################################

We also examine the `head' of the DeepMask model, focusing on score prediction. Our goal is to simplify the head and further improve inference speed.

In DeepMask, the mask and scoring heads branch after the final $512\times14\times14$ feature map (see Figure~\ref{fig:head}a). Both mask and score prediction require a large convolution, and in addition, the score branch requires an extra pooling step and hence interleaving to match the stride of the mask network during inference. Overall, this leads to a fairly inelegant and slow inference procedure.

We propose a sequence of simplified network structures that have identical mask branches but that share progressively more computation. A series of model heads A-C is detailed in Figure~\ref{fig:head}. Head A removes the need for interleaving in DeepMask by removing max pooling and replacing the $512\times7\times7$ convolutions by $128\times10\times10$ convolutions; overall this network is much faster. Head B simplifies this by having the $128\times10\times10$ features shared by both the mask and score branch. Finally, model C further reduces computation by having the score prediction utilize the same low rank $512\times1\times1$ features used for the mask.

In \S\ref{sec:exp:opt} we evaluate these variants in terms of performance and speed.

%##################################################################################################
\newcommand{\incg}[1]{\includegraphics[height=.25\textwidth]{figures/SharpMask/#1.jpg}}
\begin{figure}[!th]\centering
  \incg{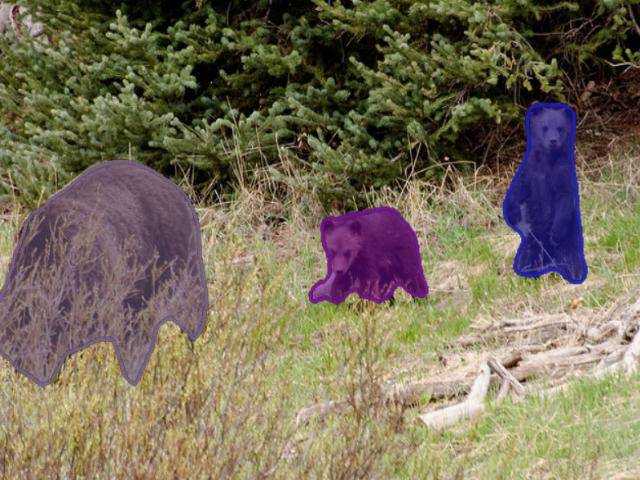}\incs\incg{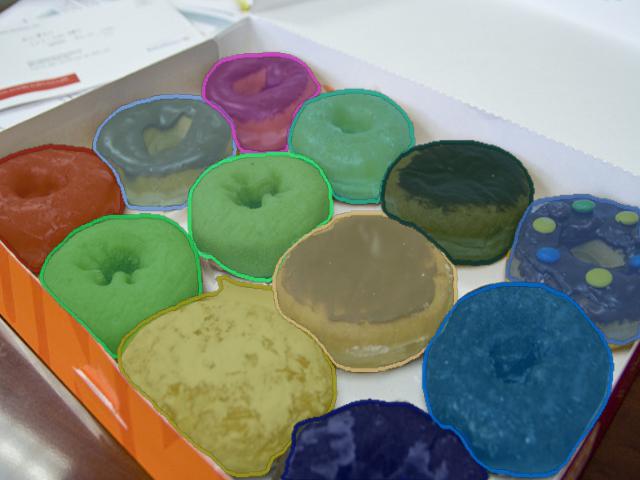}\incs\incg{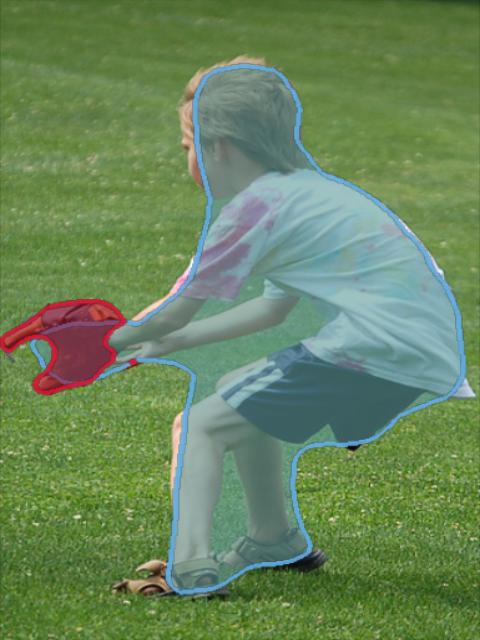}\\
  \incg{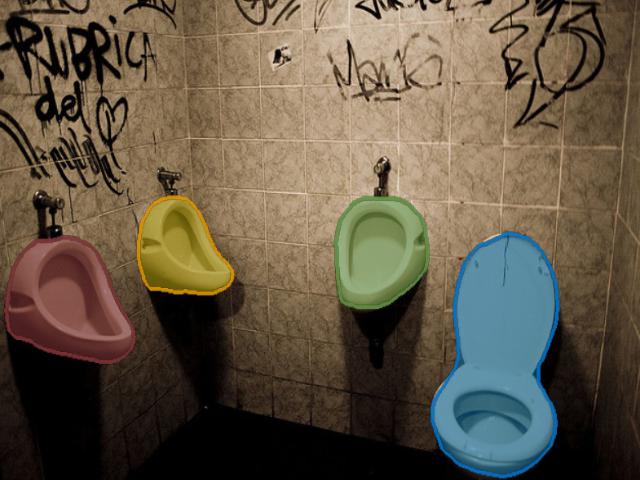}\incs\incg{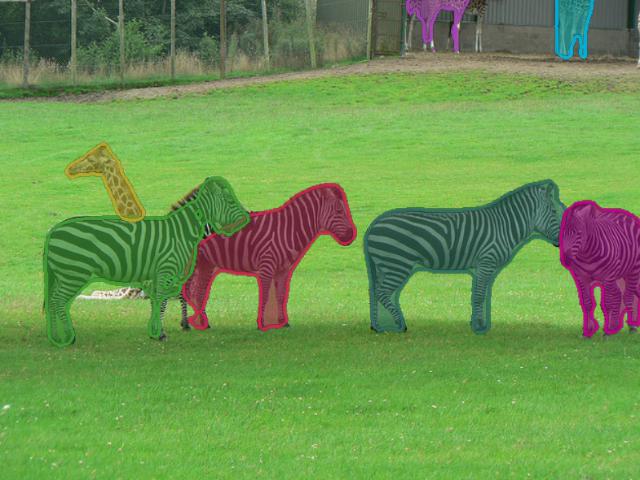}\incs\incg{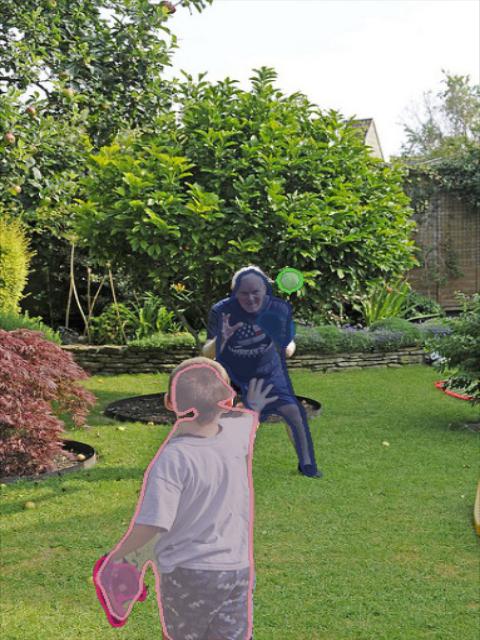}\\
  \incg{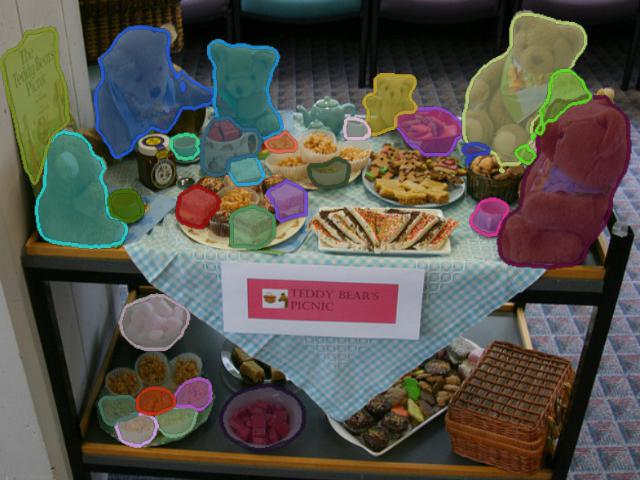}\incs\incg{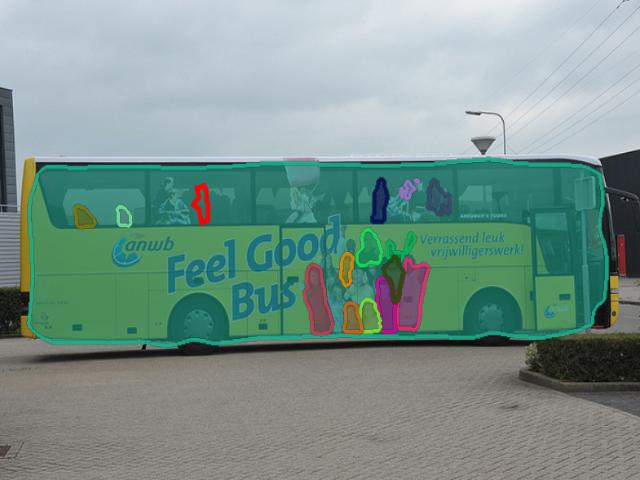}\incs\incg{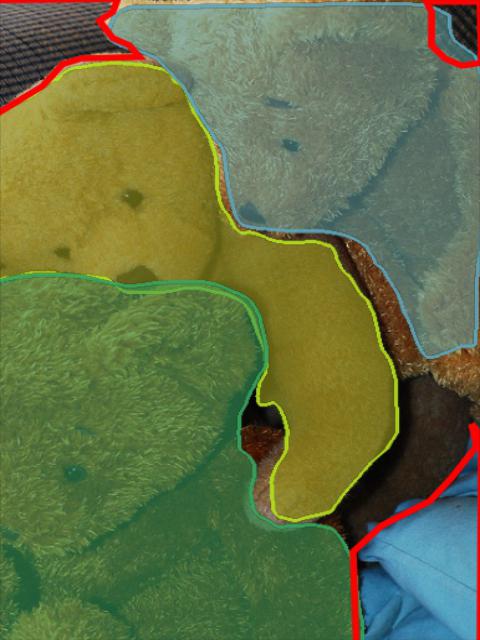}\\
  \incg{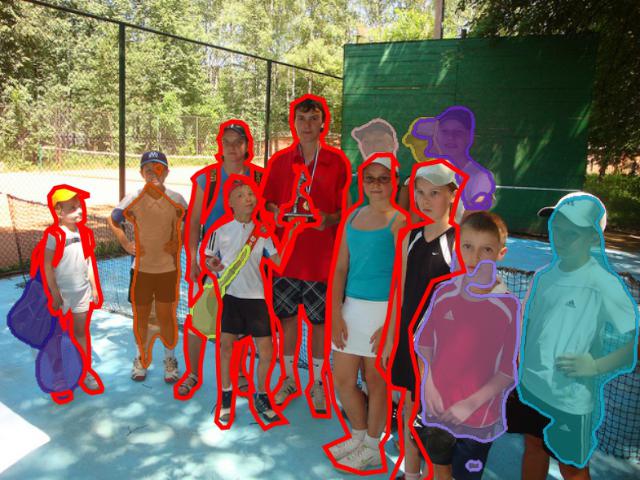}\incs\incg{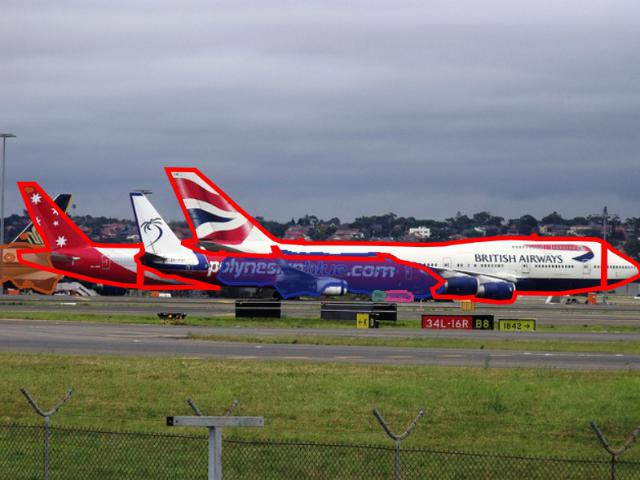}\incs\incg{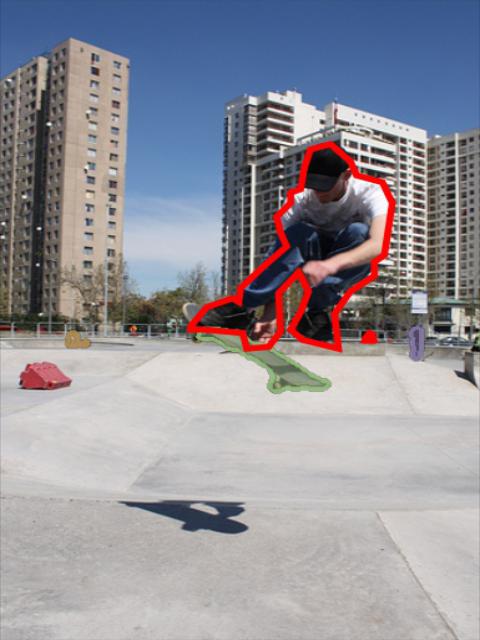}\\
\caption{SharpMask proposals with highest IoU to the ground truth on selected COCO images. Missed objects (no matching proposals with IoU $> 0.5$) are marked in red. The last row shows a number of failure cases.}
\label{fig:qualitative}
\end{figure}
%##################################################################################################

%%%%%%%%%%%%%%%%%%%%%%%%%%%%%%%%%%%%%%%%%%%%%%%%%%%%%%%%%%%%%%%%%%%%%%%%%%%%%%%%%%%%%%%%%%%%%%%%%%%
\section{Experiments}\label{sec:exp}

We train our model on the training set of the COCO dataset~\cite{mscoco2015}, which contains 80k training images and 500k instance annotations. For most of our experiments, results are reported on the first 5k COCO validation images. Mask accuracy is measured by Intersection over Union (IoU) which is the ratio of the intersection of the predicted mask and ground truth annotation to their union. A common method for summarizing object proposal accuracy is using the average recall (AR) between IoU 0.5 and .95 for a fixed number of proposals. Hosang et al.~\cite{Hosang2015proposals} show that AR correlates well with object detector performance.

Our results are measured in terms of AR at 10, 100, and 1000 proposals and averaged across all counts (AUC). As the COCO dataset contains objects in a wide range of scales, it is also common practice to divide objects into roughly equally sized sets according to object pixel area $a$: small ($a<32^2$), medium ($32^2\le a\le96^2$), and large ($a>96^2$) objects, and report accuracy at each scale.

We use a different subset of the COCO validation set to decide architecture choices and hyper-parameter selection. We use a learning rate of 1e-3 for training the refinement stage, which takes about 2 days to train on an Nvidia Tesla K40m GPU. To mitigate the mismatch of per-patch training with convolutional inference, we found that training deeper model such as ResNet requires adding extra image content (32 pixels) surrounding the training patches and using reflective-padding instead of 0-padding at every convolutional layer. Finally, following~\cite{pinheiro2015learning}, we binarize our continuous mask prediction using a threshold of 0.2.

%##################################################################################################
\begin{table}[t]\centering\scriptsize
\renewcommand\arraystretch{1.1}\renewcommand{\tabcolsep}{1.5mm}
\begin{tabular}{ l | c c c c c c | c c c c | r }
   & W & P & D & S & kernel & F & AR & AR\tss{S} & AR\tss{M} & AR\tss{L} & time \\
\shline
DeepMask          & 224 & 4  & 8  & 14 & 512x14x14  & 512 & 36.6 & 18.2 & 48.7 & 50.6 & 1.32s\\
W160-P4-D8-VGG    & 160 & 4  & 8  & 10 & 1024x10x10 & 512 & 35.5 & 15.1 & 47.5 & 53.2 &  .58s\\
\hline
W160-P4-D39       & 160 & 4 & 39  & 10 & 1024x10x10 & 512 & 37.0 & 15.9 & 50.5 & 53.9 &  .58s\\
W160-P4-D39-F128  & 160 & 4 & 39  & 10 & 1024x10x10 & 128 & 36.9 & 15.6 & 49.9 & 54.8 &  .45s\\
\hline
W112-P4-D39       & 112 & 4 & 39  & 7  & 1024x7x7   & 512 & 30.8 & 11.2 & 42.3 & 47.8 &  .31s\\
W112-P3-D21       & 112 & 3 & 21  & 14 & 512x14x14  & 512 & 36.7 & 16.7 & 49.1 & 53.1 &  .75s\\
W112-P3-D21-F128  & 112 & 3 & 21  & 14 & 512x14x14  & 128 & 36.1 & 16.3 & 48.4 & 52.2 &  .33s\\
\hline
\bf{SharpMask}    & 160 & 4 & 39  & 10 & 1024x10x10 & 128 & 39.3 & 18.1 & 52.1 & 57.1 &  .75s\\
\end{tabular}
\caption{Model performance (upper bound on AR) for varying input size W, number of pooling layers P, stride density S, depth D, and features channels F. See \S\ref{sec:trunk} and \S\ref{sec:exp:opt} for details. Timing is for multiscale inference excluding the time for score prediction. Total time for DeepMask \& SharpMask is 1.59s \& .76s.}
\label{tab:trunk}
\end{table}
%##################################################################################################

\subsection{Architecture Optimization}\label{sec:exp:opt}

We begin by reporting our optimizations of the feedforward model. For our initial results, we measure AR for densely computed masks ($\app10^4$ proposals per image). This allows us to factor out the effect of objectness score prediction and focus exclusively on evaluating mask quality. In our experiments, AR across all proposals is highly correlated (see Figure~\ref{fig:plots:nips}), hence this upper bound on AR is predictive of performance at more realistic settings (e.g. at AR\tss{100}).

\textbf{Trunk Architecture:} We begin by investigating effect of the network trunk parameters described in \S\ref{sec:trunk} with the goal of optimizing both speed and accuracy. Performance of a number of representative models is shown in Table~\ref{tab:trunk}. First, replacing the $224\times224$ DeepMask VGG-A model with a $160\times160$ version is much faster (over $2\times$). Surprisingly, accuracy loss for this model, W160-P4-D8-VGG, is only minor, partially due to an improved learning schedule. Upgrading to a ResNet trunk, W160-P4-D39, restores accuracy and keeps speed identical. We found that reducing the feature dimension to 128 (-F128) shows almost no loss, but improves speed. Finally, as input size is a bottleneck, we also tested a number of W112 models. Nevertheless, overall, W160-P4-D39-F128 gave the best tradeoff between speed and accuracy.

\textbf{Head Architecture:} In Table~\ref{tab:head} we evaluate the performance of the various network heads in Figure~\ref{fig:head} (using standard AR, not upper-bound AR as in Table~\ref{tab:trunk}). Head A is already substantially faster than DeepMask. All heads achieve similar accuracy with a decreasing inference time as the score branch shares progressively more computation with the mask. Interestingly, head C is able to predict both the score and mask from a single compact 512 dimensional vector. We chose this variant due to its simplicity and speed.

\textbf{DeepMask-ours:} Based on all of these observations, we combine the W160-P4-D39-F128 trunk with the C head. We refer to the resulting architecture as \emph{DeepMask-ours}. DeepMask-ours is over $3\times$ faster than the original DeepMask (.46s per image versus 1.59s) and also more accurate. Moreover, model parameter count is reduced from \app75M to \app17M. For all SharpMask experiments, we adopt DeepMask-ours as the base feedforward architecture.

%##################################################################################################
\begin{table}[t]\centering\scriptsize
\renewcommand\arraystretch{1.1}\renewcommand{\tabcolsep}{1.5mm}
\begin{tabular}{c|lllllll|rrr}
  & AR$^{10}$ & AR$^{100}$ & AR$^{1K}$ & AUC\tss{S} & AUC\tss{M} & AUC\tss{L} & AUC
  & mask & score & total \\
\shline
  DeepMask & 12.6 & 24.5 & 33.1 & 2.3 & 26.6 & 33.6 & 18.3 & 1.32s & .27s & 1.59s\\
\hline
  head A  & 14.0 & 25.8 & 33.4 & 2.2 & 27.3 & 36.6 & 19.3 &  .45s & .06s &  .51s\\
  head B  & 14.0 & 25.4 & 33.0 & 2.0 & 27.0 & 36.9 & 19.1 &  .45s & .05s &  .50s\\
  head C  & 14.4 & 25.8 & 33.1 & 2.2 & 27.3 & 37.4 & 19.4 &  .45s & .01s &  .46s\\
\end{tabular}
\caption{All model variants of the head have similar performance. Head C is a win in terms of both simplicity and speed. See Figure~\ref{fig:head} for head definitions.}
\label{tab:head}
\end{table}
%##################################################################################################

\subsection{SharpMask Analysis}\label{sec:exp:sharp}

%##################################################################################################
\begin{figure}[t]\centering
  \begin{subfigure}[b]{0.32\textwidth}
    \centering\includegraphics[width=\textwidth]{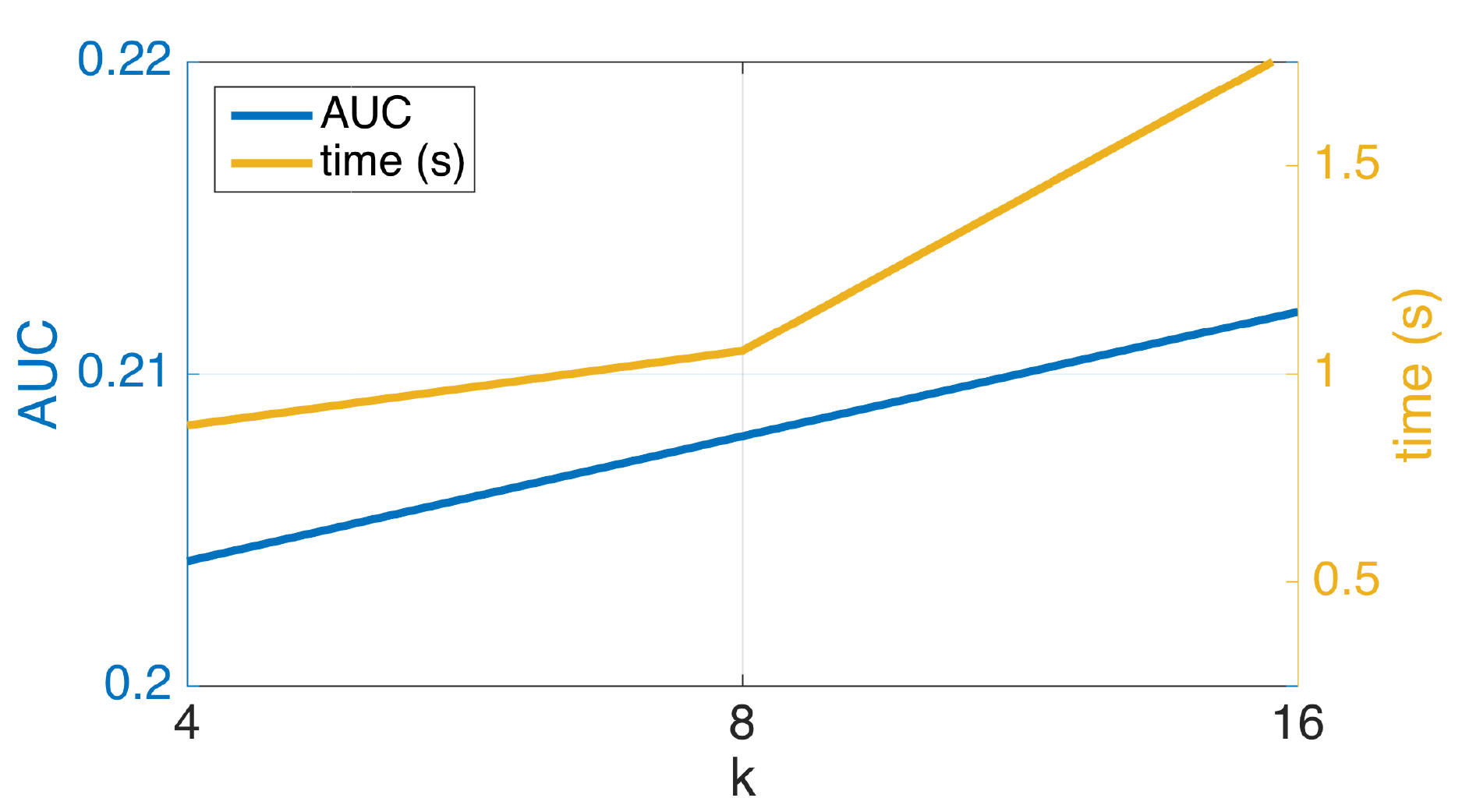}
    \caption{$k_m^i=k_s^i=k$}
  \end{subfigure}\hspace{1mm}
  \begin{subfigure}[b]{0.32\textwidth}
    \centering\includegraphics[width=\textwidth]{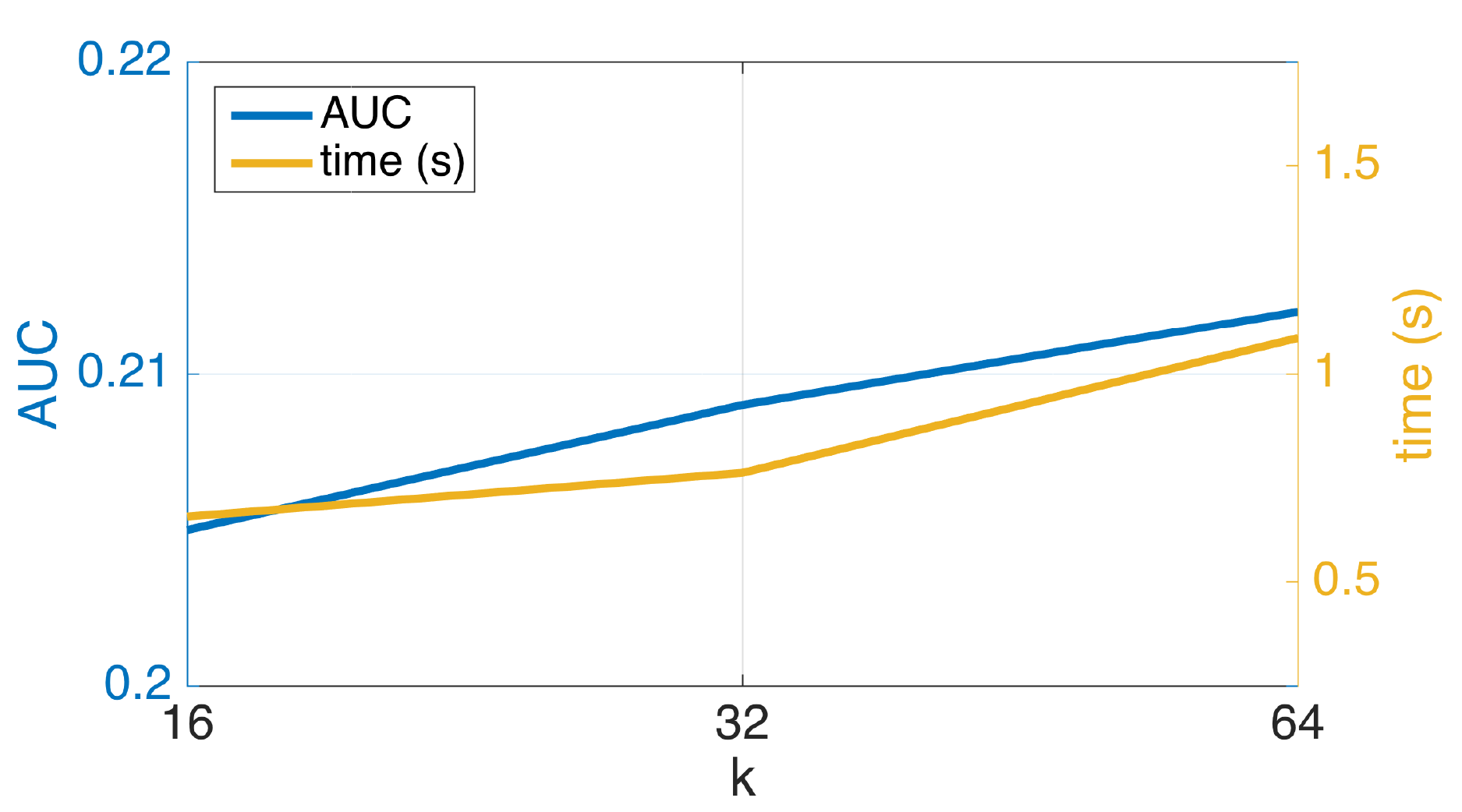}
    \caption{$k_m^i=k_s^i=\frac{k}{2^{i-1}}$}
  \end{subfigure}\hspace{1mm}
  \begin{subfigure}[b]{0.3\textwidth}
    \centering\includegraphics[width=\textwidth]{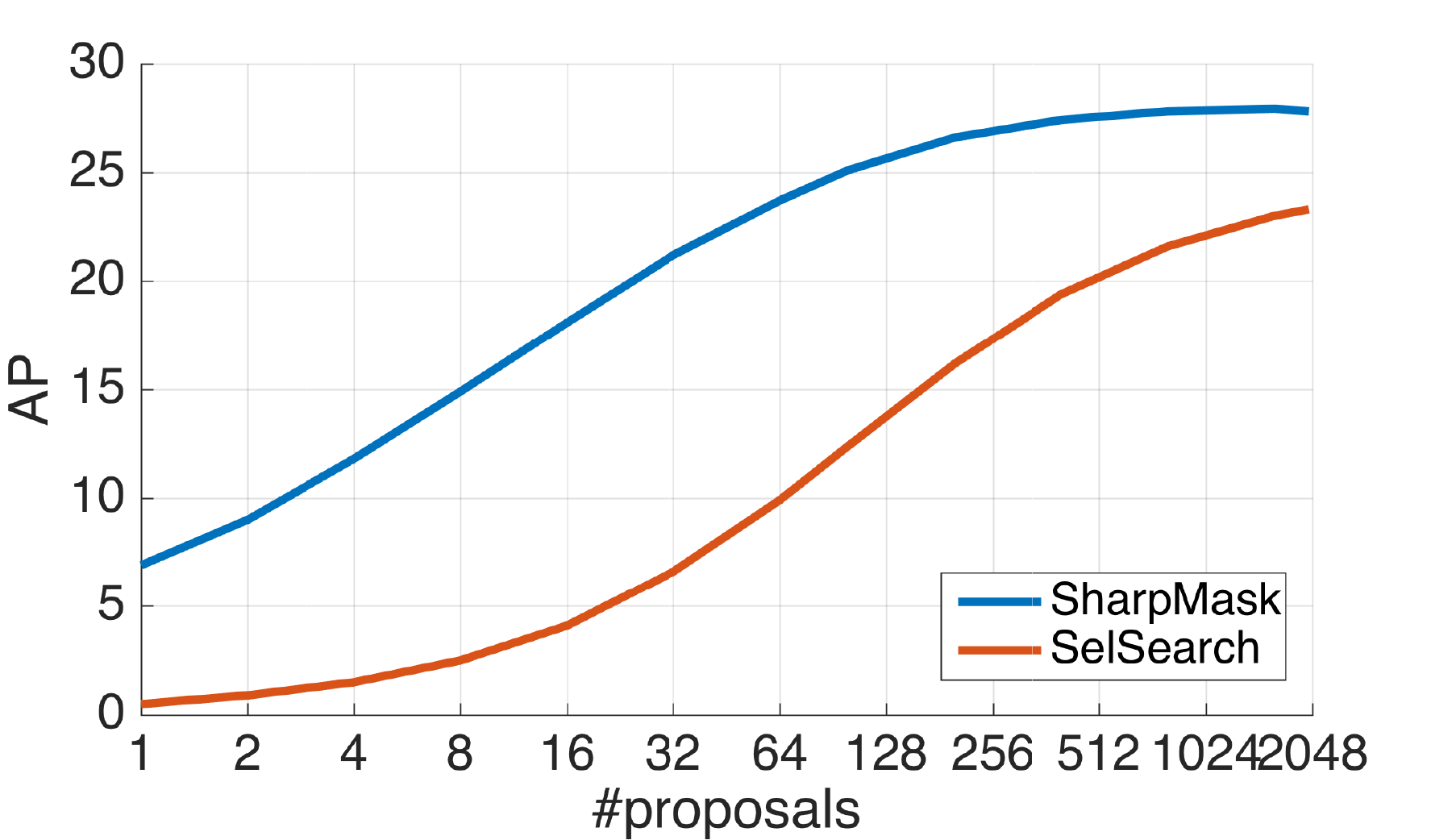}
    \caption{detection perf. }\label{fig:detection}
  \end{subfigure}
\caption{(a-b) Performance and inference time for multiple SharpMask variants. (c) Fast R-CNN detection performance versus number and type of proposals.}
\label{fig:plots}
\end{figure}
%##################################################################################################

We now analyze different parameter settings for our top-down refinement network. As described in \S\ref{sec:refinement}, each of the four refinement modules $R^i$ in SharpMask is controlled by two parameters $k_m^i$ and $k_s^i$, which denote the size of the mask encoding $M^i$ and skip encoding $S^i$, respectively. These parameters control network capacity and effect inference speed. We experiment with two different schedules for these parameters: (a) $k_m^i=k_s^i=k$ and (b) $k_m^i=k_s^i=\frac{k}{2^{i-1}}$ for each $i\le4$.

Figure~\ref{fig:plots}(a-b) shows performance for the two schedules for different $k$ both in terms of AUC and inference time (measured when refining the top 500 proposals per image, at which point object detection performance saturates, see Figure~\ref{fig:plots}c). We consistently observe higher performance as we increase the capacity, with no sign of overfitting. Parameter schedule b, in particular with $k=32$, has the best trade-off between performance and speed, so we chose this as our final model.

We note that we were unable to obtain good results with schedule a for $k\le2$, indicating the importance of using sufficiently large $k$. Also, we observed that a single $3\times3$ convolution encounters learning difficulties when ($k^i_s \ll k^i_f$). Therefore, in all experiments we used a sequence of two $3\times3$ convolutions (followed by ReLUs) to generate $S^i$ from $F^i$, reducing $F^i$ to 64 channels first followed by a further reduction to $k^i_s$ channels.

Finally, we performed two additional ablation studies. First, we removed all downward convs, set $k_m^i=k_s^i=1$, and averaged the output of all layers. Second, we kept the vertical convs but removed all horizontal convs. These two variants are related to `skip' and `deconv' networks, respectively. Neither setup showed meaningful improvement over the baseline feedforward network. In short, we found that both horizontal and vertical connections were necessary for this task.

\subsection{Comparison with State of the Art}\label{sec:exp:comparison}

%##################################################################################################
\begin{table}[t]\centering\scriptsize
\renewcommand\arraystretch{1.1}\renewcommand{\tabcolsep}{.5mm}
\begin{tabular}{l@{\hskip 2mm}|cccc@{\hskip 3mm}|cccccccc}
 & \multicolumn{4}{c|}{\bf Box Proposals} &
 & \multicolumn{7}{c}{\bf Segmentation Proposals} \\
 & AR$^{10}$ & AR$^{100}$ & AR$^{1K}$ & AUC &
 & AR$^{10}$ & AR$^{100}$ & AR$^{1K}$ & AUC\tss{S} & AUC\tss{M} & AUC\tss{L} & AUC\\
\shline
 EdgeBoxes~\cite{ZitnickD14}              &  7.4 & 17.8 & 33.8 & 13.9 & & |    & |    & |    & |    & |    & |    & |   \\
 Geodesic~\cite{KrahenbuhlK14}            &  4.0 & 18.0 & 35.9 & 12.6 & &  2.3 & 12.3 & 25.3 &  1.3 &  8.6 & 20.5 &  8.5\\
 Rigor~\cite{humayun2014}                 &  |   & 13.3 & 33.7 & 10.1 & & |    &  9.4 & 25.3 &  2.2 &  6.0 & 17.8 &  7.4\\
 SelectiveSearch~\cite{Uijlings13}        &  5.2 & 16.3 & 35.7 & 12.6 & &  2.5 &  9.5 & 23.0 &  0.6 &  5.5 & 21.4 &  7.4\\
 MCG~\cite{ARXIV2015MCG}                  & 10.1 & 24.6 & 39.8 & 18.0 & &  7.7 & 18.6 & 29.9 &  3.1 & 12.9 & 32.4 & 13.7\\
\hline
 RPN~\cite{bell15ion,RenNIPS15fasterRCNN} & 12.8 & 29.2 & 42.6 & 21.4 & & |    & |    & |    & |    & |    & |    & |   \\
 DeepMask~\cite{pinheiro2015learning}     & 15.3 & 31.3 & 44.6 & 23.3 & & 12.6 & 24.5 & 33.1 &  2.3 & 26.6 & 33.6 & 18.3\\
 DeepMaskZoom~\cite{pinheiro2015learning} & 15.0 & 32.6 & 48.2 & 24.2 & & 12.7 & 26.1 & 36.6 &  6.8 & 26.3 & 30.8 & 19.4\\
\hline
 DeepMask-ours                            & 18.7 & 34.9 & 46.5 & 26.2 & & 14.4 & 25.8 & 33.1 &  2.2 & 27.3 & 37.4 & 19.4\\
 SharpMask                                & 19.7 & 36.4 & 48.2 & 27.4 & & 15.6 & 27.6 & 35.5 &  2.5 & 29.1 & 40.4 & 20.9\\
 SharpMaskZoom                            & 20.1 & 39.4 & 52.8 & 29.1 & & 16.1 & 30.3 & 39.2 &  6.9 & 29.7 & 38.4 & 22.4\\
 SharpMaskZoom$^{2}$                      & 19.2 & 39.9 & 55.0 & 29.2 & & 15.4 & 30.7 & 40.8 & 10.6 & 27.3 & 36.0 & 22.5\\

\end{tabular}
\caption{Results on the COCO validation set on box and segmentation proposals. AR at different proposals counts is reported and also AUC (AR averaged across all proposal counts). For segmentation proposals, we also report AUC at multiple scales. SharpMask has largest for segmentation proposals and large objects.}
\label{tab:rec:coco}
\end{table}
%##################################################################################################

Table~\ref{tab:rec:coco} compares the performance of our model, SharpMask, to other existing methods on the COCO dataset. We compare results both on box and segmentation proposals (for box proposals we extract tight bounding boxes surrounding our segmentation masks). SharpMask achieves the state of the art in all metrics for both speed and accuracy by a large margin. Additionally, because SharpMask has a smaller input size, it can be applied to an additional one to two scales (\emph{SharpMaskZoom}) and achieves a large boost in AR for small objects.

Our feedforward architecture improvements, \emph{DeepMask-ours}, alone, improve over the original DeepMask, in particular for bounding box proposals. Not only is the new baseline more accurate, with our architecture optimization to the trunk and head of the network (see \S\ref{sec:architecture}), speed is improved to .46s per image. We emphasize that DeepMask was the previous state-of-the-art on this task, outperforming all bottom-up proposal methods as well as Region Proposal Networks (RPN)~\cite{RenNIPS15fasterRCNN} (we obtained improved RPN proposals from the authors of~\cite{bell15ion}). 

We train SharpMask using DeepMask-ours as the feedforward network. As the two networks have an identical score branch, we can disentangle the performance improvements achieved by our top-down refinement approach. Once again, we observe a considerable boost in performance on AR due to the top-down refinement. We note that improvement for segmentation predictions is bigger than box predictions, which is not surprising, as sharpening masks might not change the tight box around the objects in many examples. Inference for SharpMask is .76s per image, over $2\times$ faster than DeepMask; moreover, the refinement modules require fewer than 3M additional parameters.

In Figures~\ref{fig:comparison} and \ref{fig:comparison:more} we show direct comparison between SharpMask and DeepMask and we can see SharpMask generates higher-fidelity masks that more accurately delineate object boundaries. In Figures~\ref{fig:qualitative} and \ref{fig:qualitative:more}, we show more qualitative results. Additional detailed performance plots are shown in Figure~\ref{fig:plots:nips}.

\subsection{Object Detection}\label{sec:exp:detection}

In this section, we use SharpMask in the Fast R-CNN pipeline~\cite{girshick15fastrcnn} and analyze the improvements of using our proposals for object detection. In the following experiments we coupled SharpMask proposals with two classifiers: VGG~\cite{Simonyan15} and MultiPathNet (MPN)~\cite{zagoruyko2016multipath}, which introduces a number of improvements to the VGG classifier. In future work we will also test our proposals with ResNets~\cite{he2015deep}. 

First, Fig.~\ref{fig:detection} shows the comparison of bounding box detection results for SharpMask and SelSearch~\cite{Uijlings13} on the COCO val set with the MPN classifier applied to both. SharpMask achieves 28 AP, which is 5 AP higher than SelSearch. Also, performance converges using only \app500 SharpMask proposals per image.

Next, Table~\ref{tab:leaderboard} top shows results of various baselines without bells and whistles, trained on the train set only. SharpMask achieves top results with the VGG classifier, outperforming both RPN~\cite{RenNIPS15fasterRCNN} and SelSearch~\cite{Uijlings13}.

Finally, Table~\ref{tab:leaderboard} middle/bottom shows results from the 2015 COCO detection challenges. The performance is reported with model ensembling and the MPN classifier. The ensemble model achieve 33.5 AP for boxes and 25.1 AP for segments, and achieved second place in the challenges. Note that for the challenges, both SharpMask and MPN used the VGG trunk (ResNets were concurrent work, and won the competitions). We have not re-run our model with ensembling and additional bells and whistles after integrating ResNets into SharpMask.

%##################################################################################################
\begin{table}[t]\centering\scriptsize
\renewcommand\arraystretch{1.1}\renewcommand{\tabcolsep}{.8mm}
\begin{tabular}{l|cccccccccccc}
  & AP & AP\tss{50} & AP\tss{75} & AP\tss{S} & AP\tss{M} & AP\tss{L} &
  AR\tss{1} & AR\tss{10} & AR\tss{100} & AR\tss{S} & AR\tss{M} & AR\tss{L} \\
 \shline
    SelSearch + VGG \cite{girshick15fastrcnn}
    & 19.3 & 39.3 & | & | & | & | & | & | & | & | & | & |\\
   RPN + VGG \cite{RenNIPS15fasterRCNN}
    & 21.9 & 42.7 & | & | & | & | & | & | & | & | & | & |\\
   SharpMask + VGG
    & 25.2 & 43.4 & | & | & | & | & | & | & | & | & | & |\\
 \hline    
   ResNet++ \cite{he2015deep} 
    & 28.2 & 51.5 & 27.9 & 9.3 & 30.6 & 45.2 & 25.7 & 37.4 & 38.2 & 16.8 & 43.9 & 57.6\\
   SharpMask+MPN\cite{zagoruyko2016multipath} 
    & 25.1 & 45.8 & 24.8 & 7.4 & 29.2 & 39.1 & 24.1 & 36.8 & 38.7 & 17.3 & 46.9 & 53.9\\
 \hline
   ResNet++ \cite{he2015deep}
    & 37.3 & 58.9 & 39.9 & 18.3 & 41.9 & 52.4 & 32.1 & 47.7 & 49.1 & 27.3 & 55.6 & 67.9\\
   SharpMask+MPN\cite{zagoruyko2016multipath}
    & 33.5 & 52.6 & 36.6 & 13.9 & 37.8 & 47.7 & 30.2 & 46.2 & 48.5 & 24.1 & 56.1 & 66.4\\
   ION \cite{bell15ion}
    & 31.0 & 53.3 & 31.8 & 12.3 & 33.2 & 44.7 & 27.9 & 43.1 & 45.7 & 23.8 & 50.4 & 62.8\\
\end{tabular}
\caption{\textbf{Top}: COCO bounding box results of various baselines without bells and whistles, trained on the train set only, and reported on test-dev (results for~\cite{girshick15fastrcnn, RenNIPS15fasterRCNN} obtained from original papers). We denote methods using `proposal+classifier' notation for clarity. SharpMask achieves top results, outperforming both RPN and SelSearch proposals. \textbf{Middle}: Winners of the 2015 COCO segmentation challenge. \textbf{Bottom}: Winners of the 2015 COCO bounding box challenge.}
\label{tab:leaderboard}
\end{table}
%##################################################################################################

%%%%%%%%%%%%%%%%%%%%%%%%%%%%%%%%%%%%%%%%%%%%%%%%%%%%%%%%%%%%%%%%%%%%%%%%%%%%%%%%%%%%%%%%%%%%%%%%%%%
\section{Conclusion}\label{sec:conclusion}

In this paper, we introduce a novel architecture for object instance segmentation, based on an augmentation of feedforward networks with top-down refinement modules. Our model achieves a new state of the art for object proposals generation, both in terms of performance and speed. The proposed refinement approach is general and could be applied to other pixel-labeling tasks.

%##################################################################################################
\newcommand{\showplot}[2]{
  \begin{subfigure}{0.31\textwidth}
  \includegraphics[width=\textwidth]{figures/plots/#1.pdf}
  \vspace{-5mm}\caption{\scriptsize #2}\vspace{2mm}\end{subfigure}}
\begin{figure}[t]\centering
  \showplot{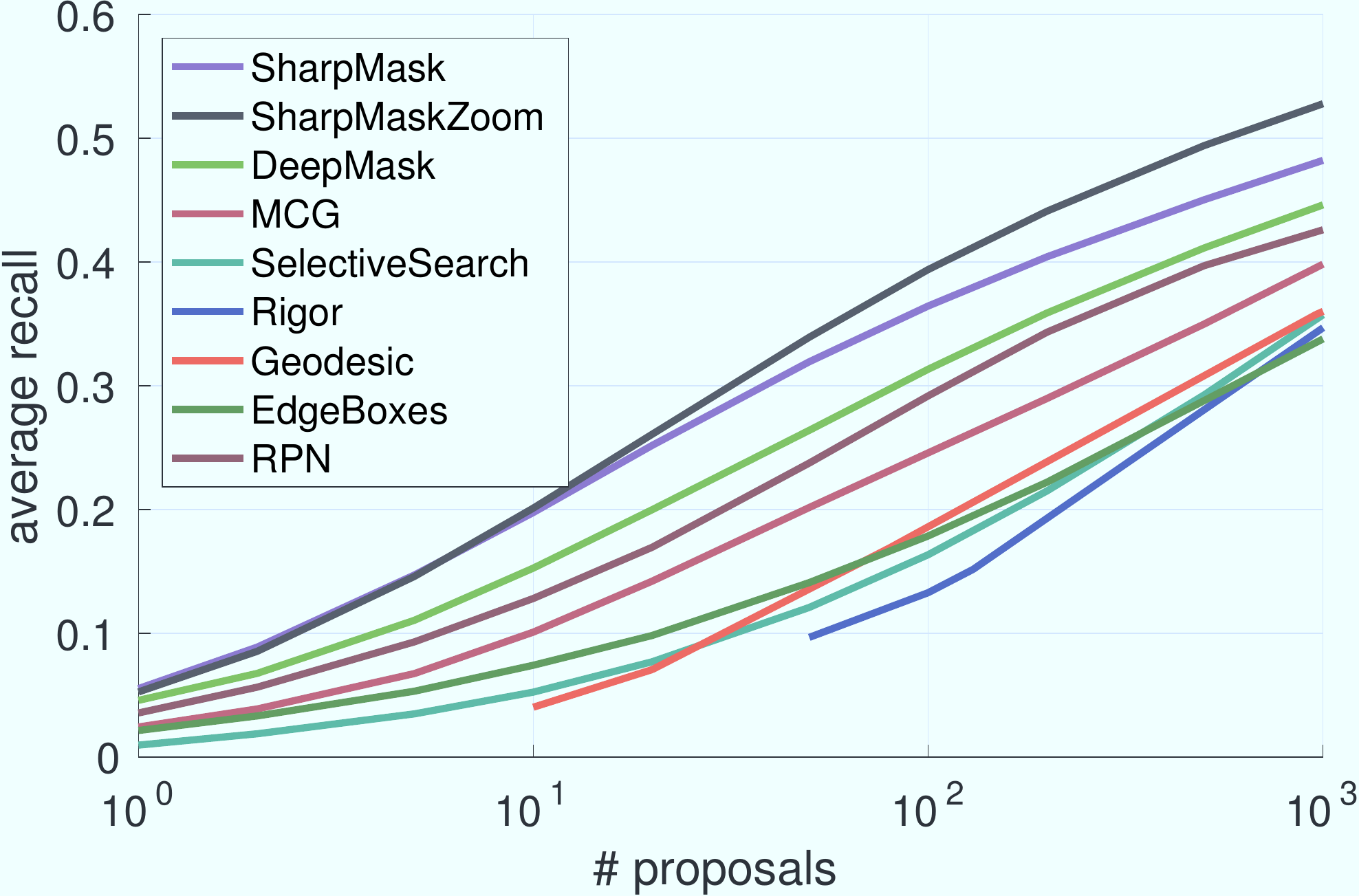}{Bounding box proposals}\hspace{5mm}
  \showplot{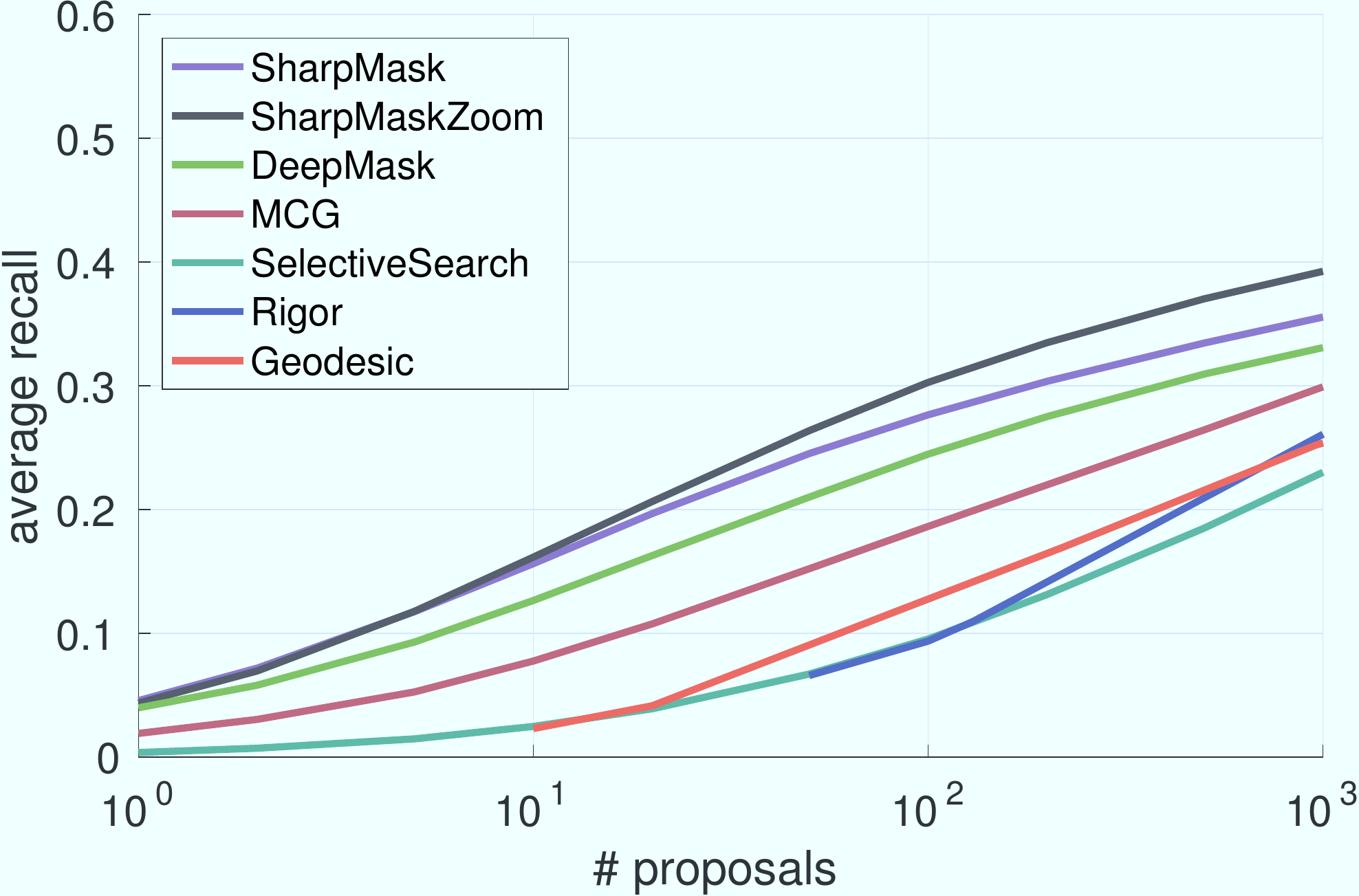}{Segmentation proposals}\\
  \showplot{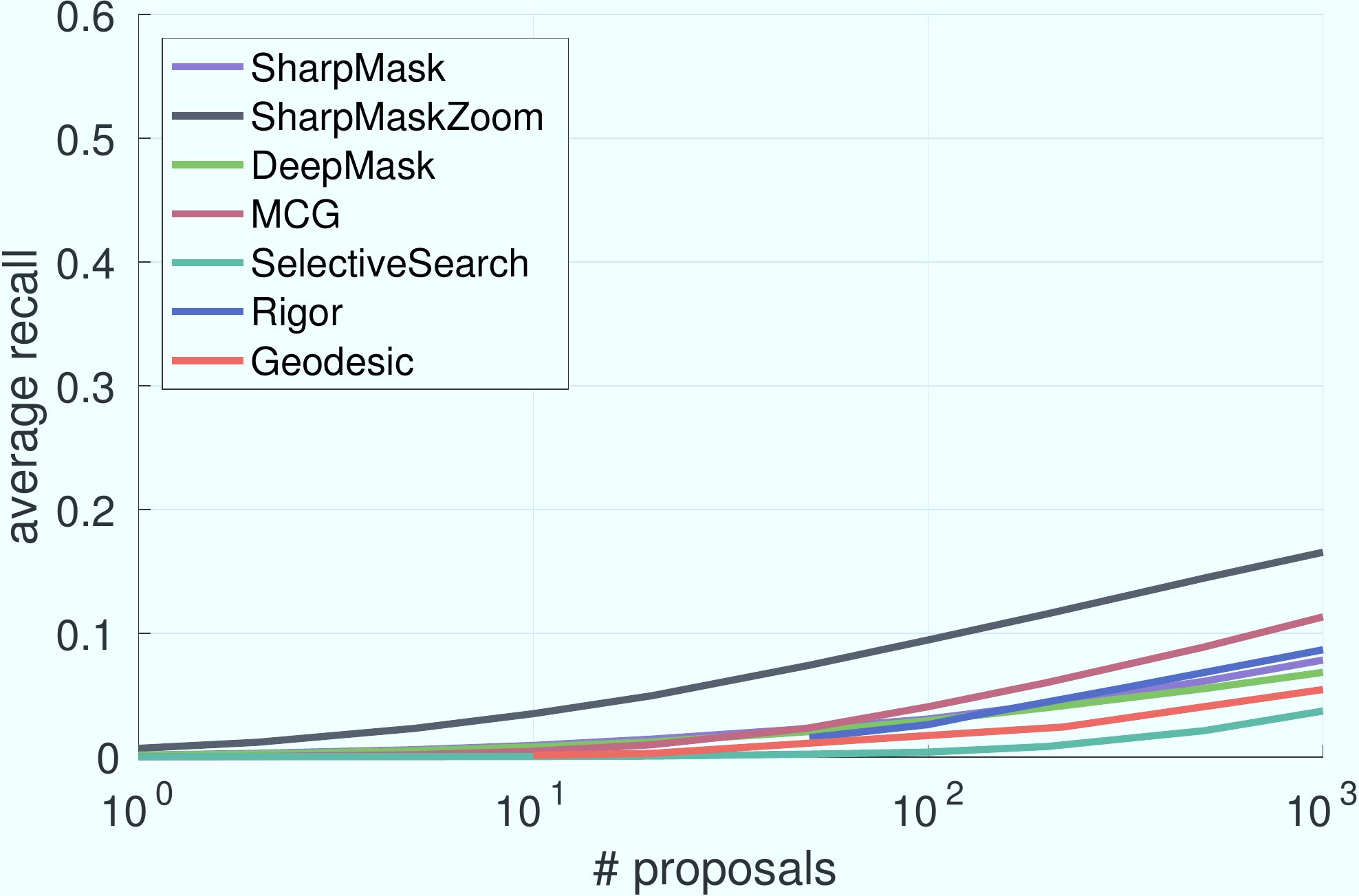}{Small objects ($a<32^2$)}%
  \showplot{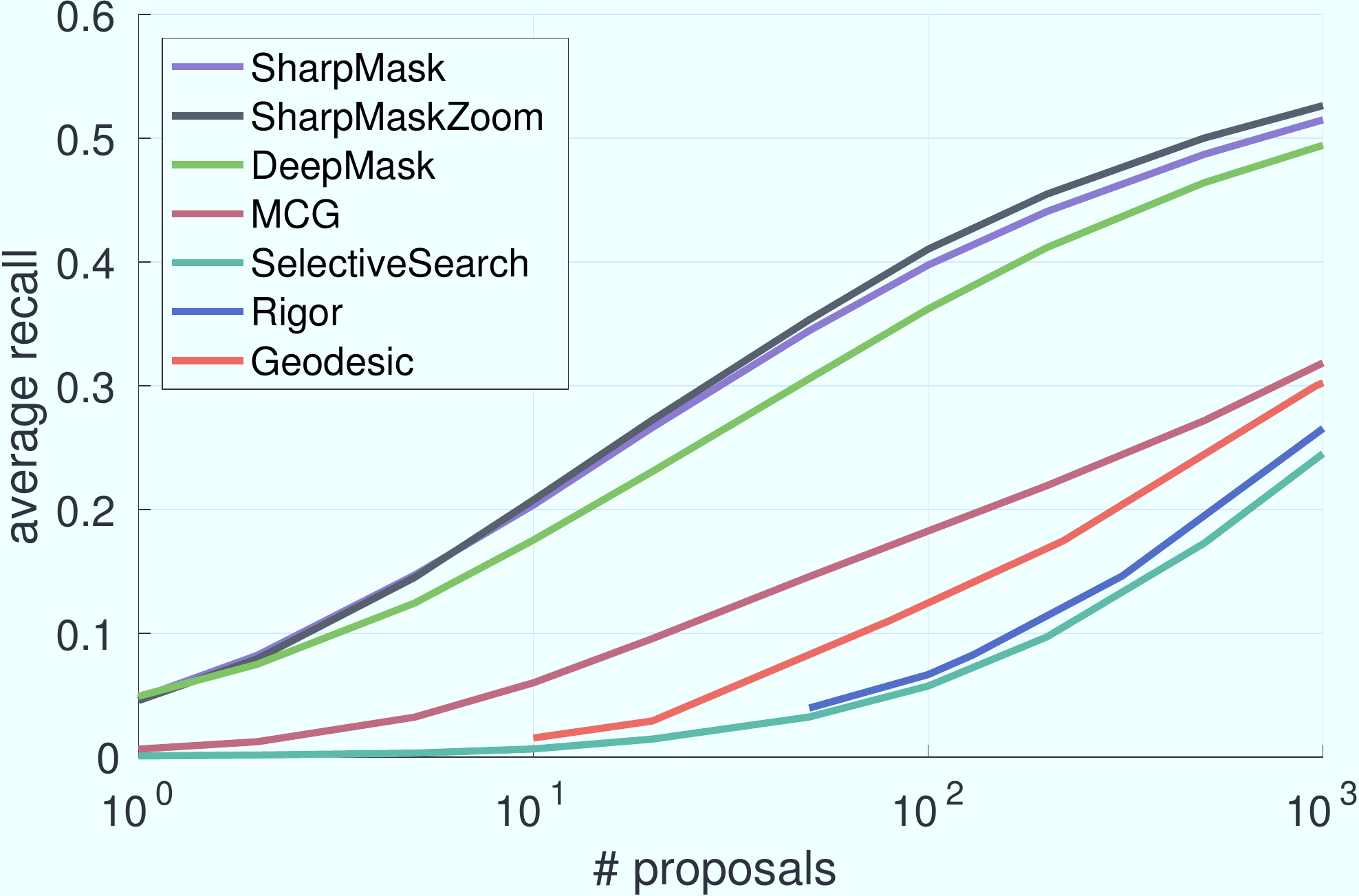}{Medium objects}%
  \showplot{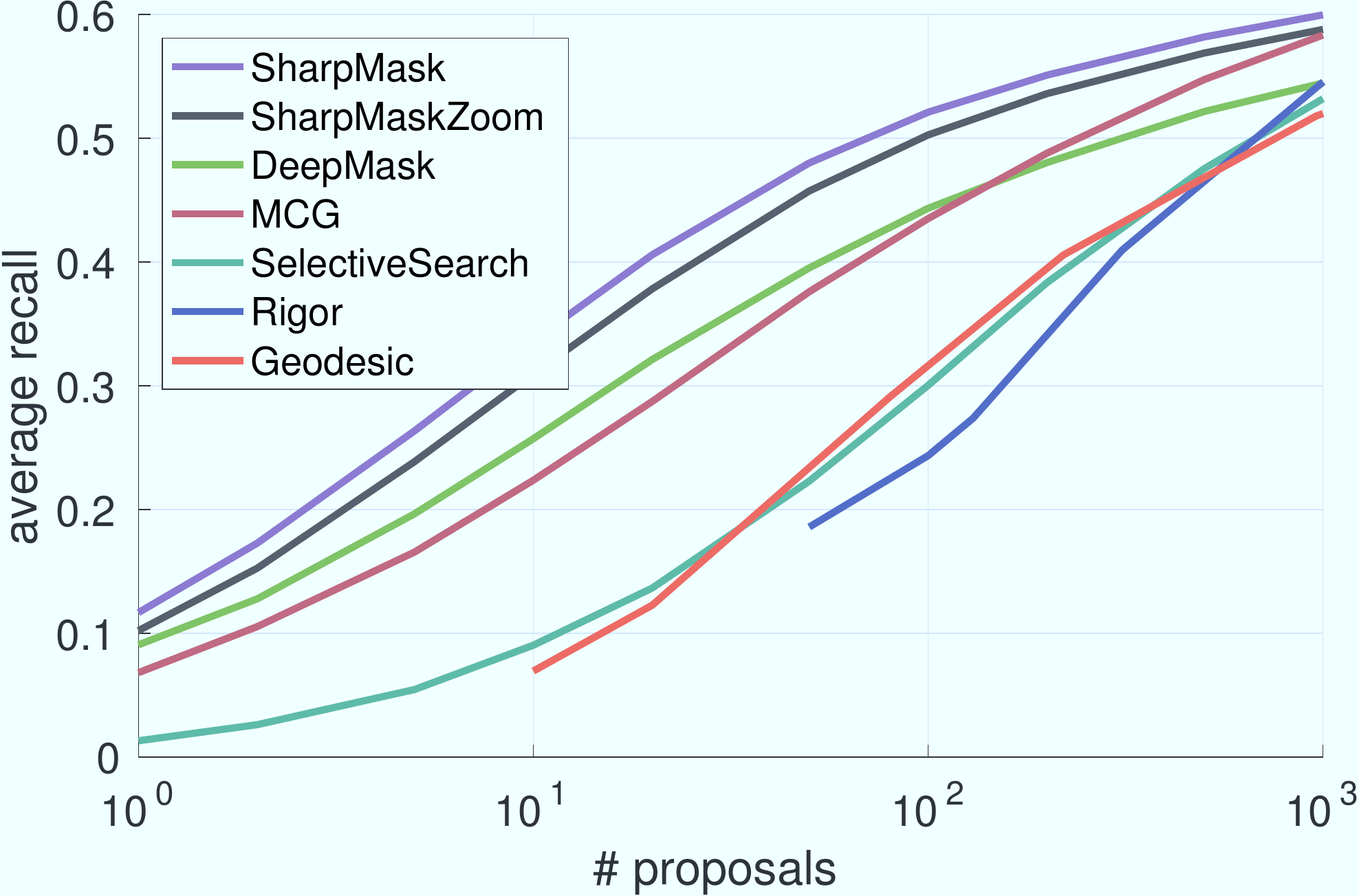}{Large objects ($a>96^2$)}\\
  \showplot{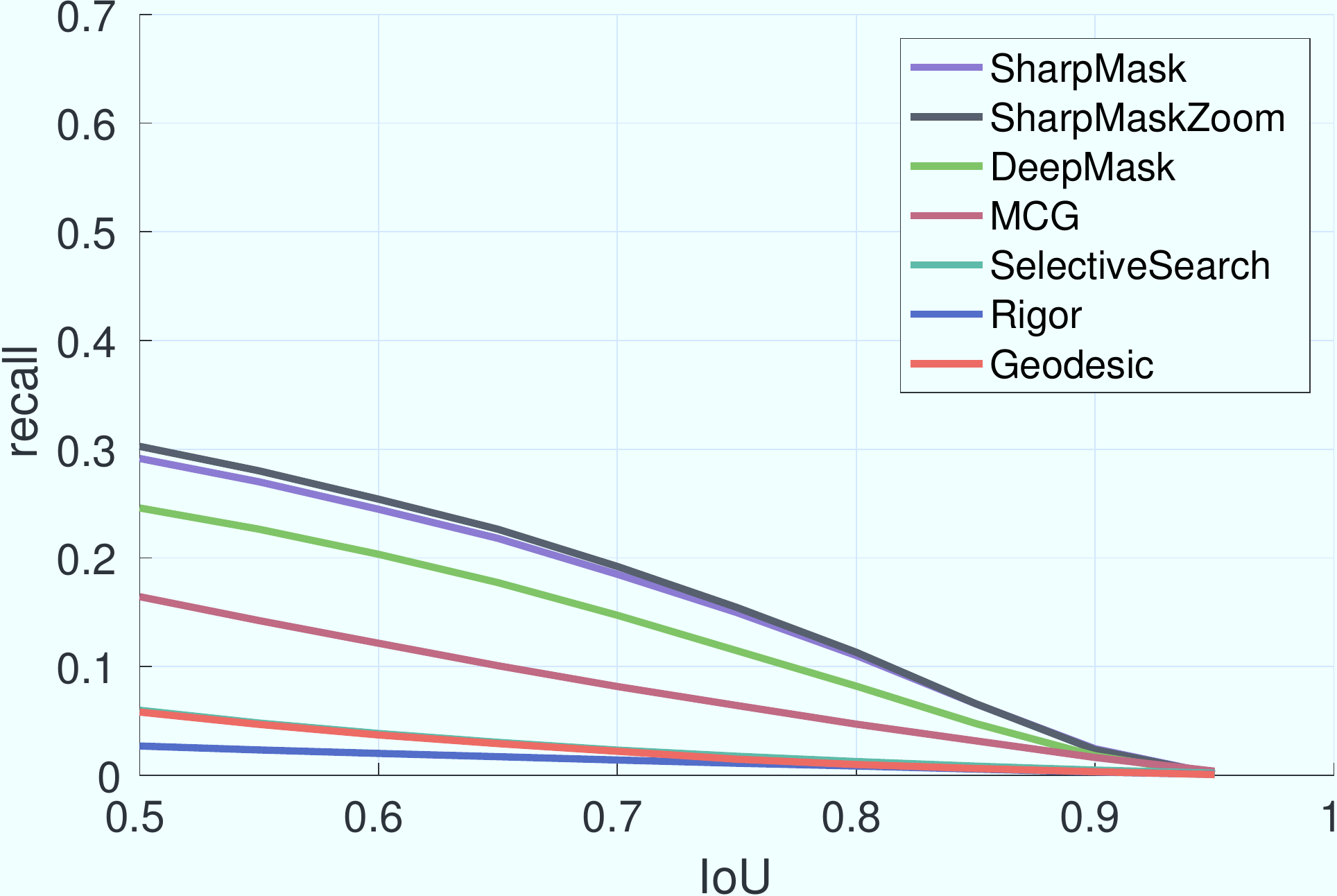}{Recall @10 proposals}%
  \showplot{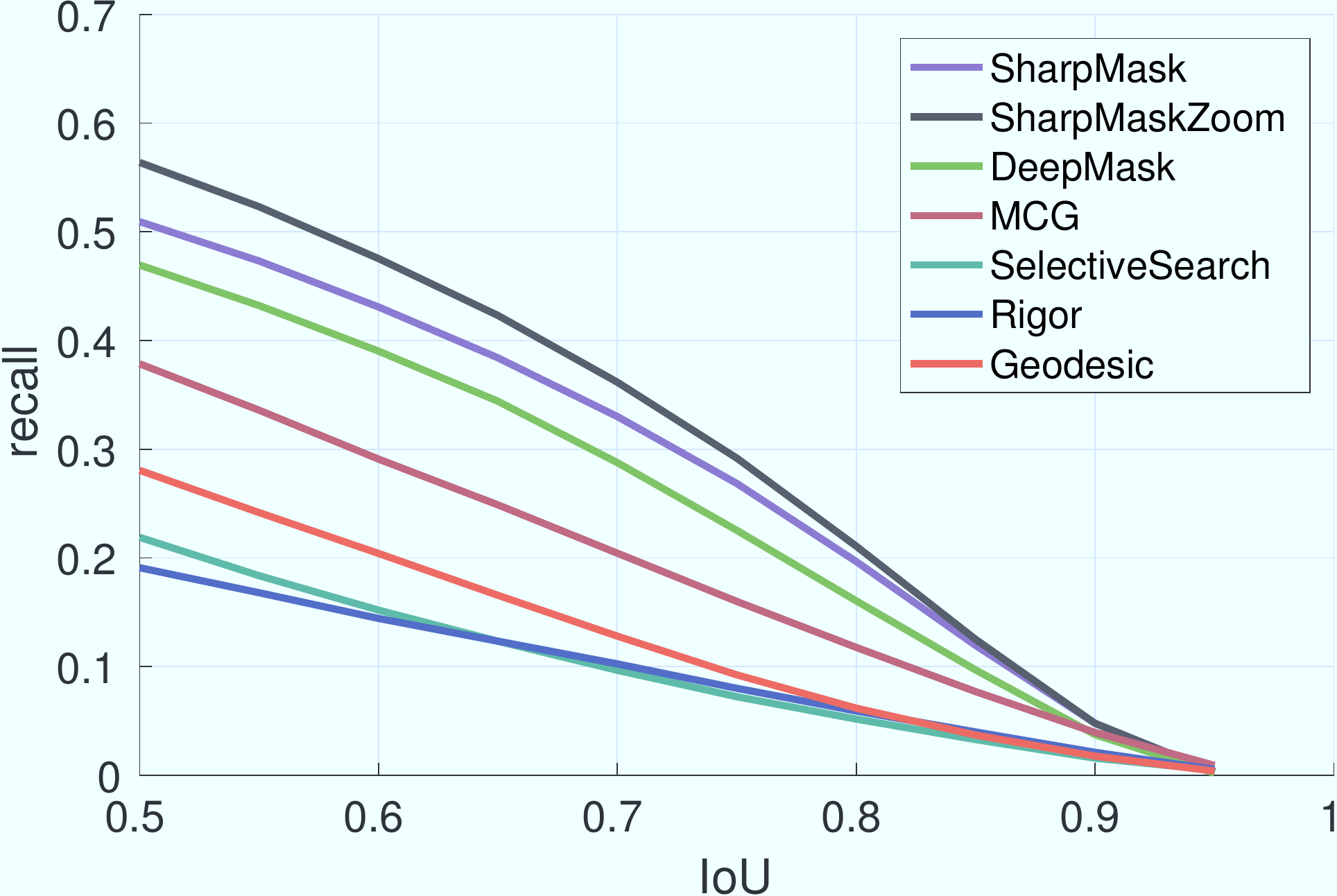}{Recall @100 proposals}%
  \showplot{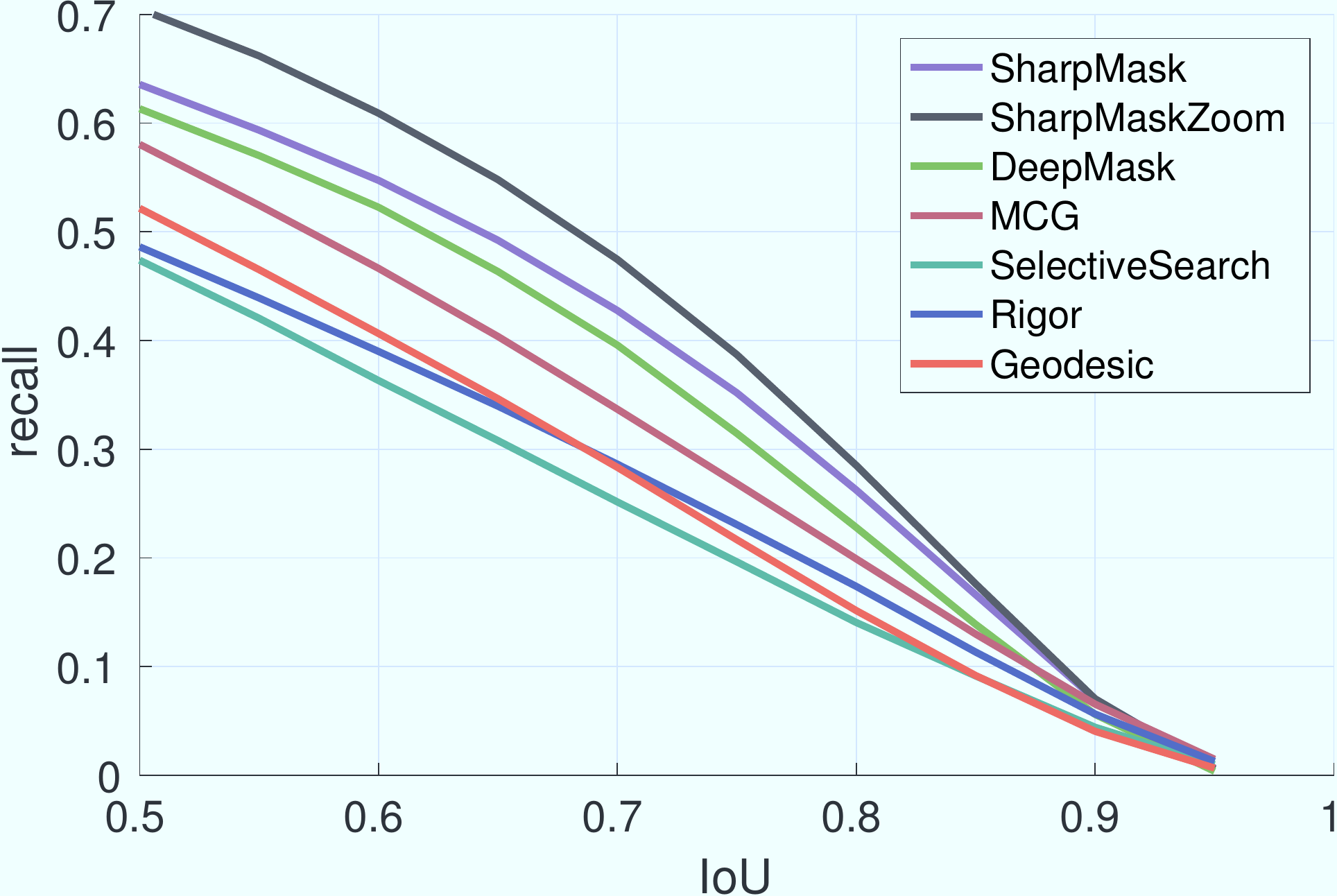}{Recall @1000 proposals}\vspace{-3mm}
  \caption{\small(a-b) Average recall versus number of box and segment proposals on COCO. (c-e) AR versus number of proposals for different object scales on segment proposals. (f-h) Recall versus IoU threshold for different number of segment proposals.}
  \label{fig:plots:nips}
\end{figure}
%##################################################################################################

%##################################################################################################
\begin{figure}[t]\centering
  \includegraphics[width=.99\textwidth]{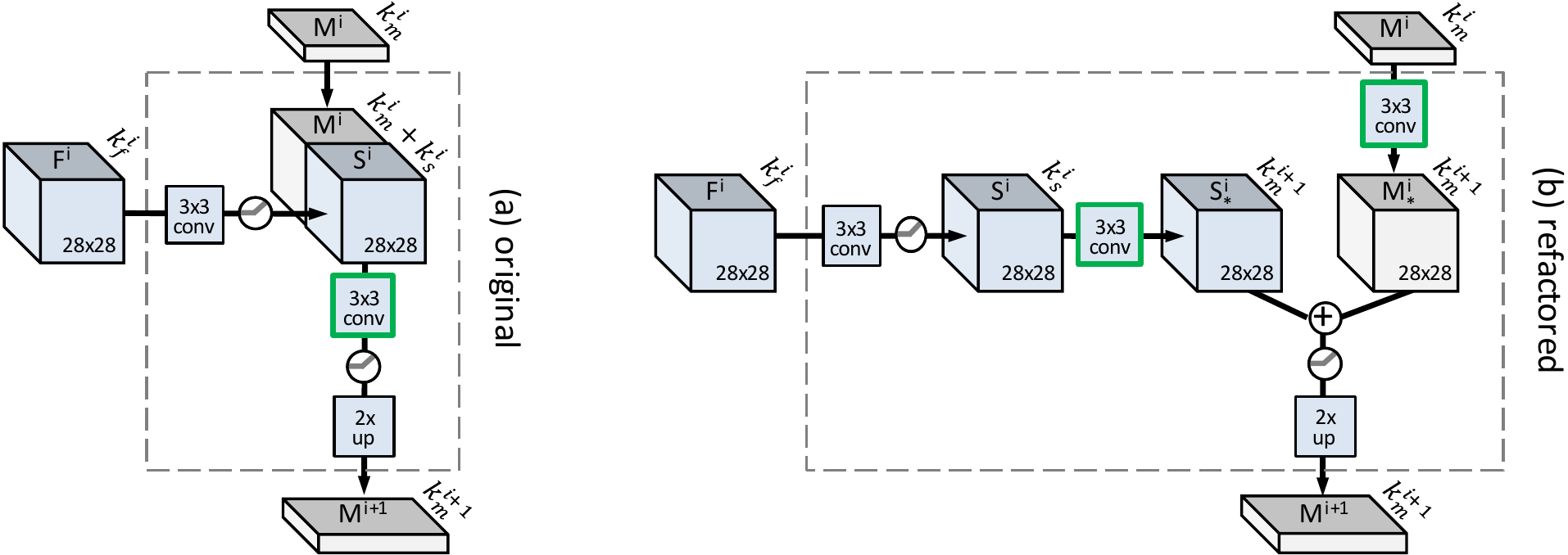}
\caption{\small(a) Original refinement model. (b) Refactored but \emph{equivalent} model that leads to a more efficient implementation. The models are equivalent as concatenating along depth and convolving along the spatial dimensions can be rewritten as two separate spatial convolutions followed by addition. The green `conv' boxes denote the corresponding convolutions (note also the placement of the ReLUs). The refactored model is more efficient as skip features (both $S^i$ and $S^i_*$) are shared by overlapping refinement windows (while $M^i$ and $M^i_*$ are not). Finally, observe that setting $k^i_m=1$, $\forall i$, and removing the top-down convolution would transform our refactored model into a standard `skip' architecture (however, using $k^i_m=1$ is not effective in our setting).}
\label{fig:equivalence}
\end{figure}
\clearpage
%##################################################################################################

%##################################################################################################
\begin{figure}[h]\centering
  \incg{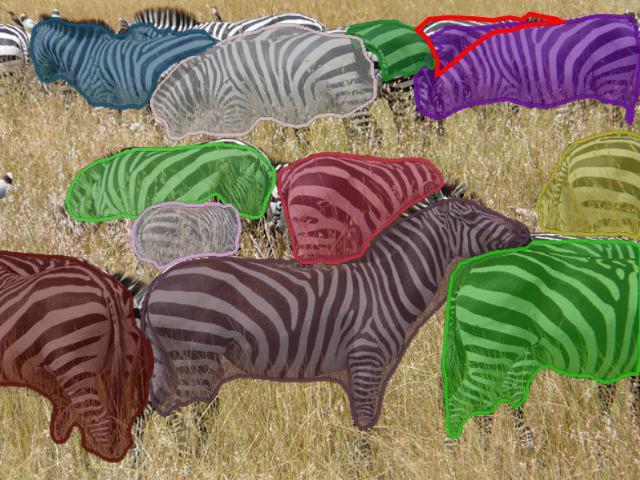}\incs\incg{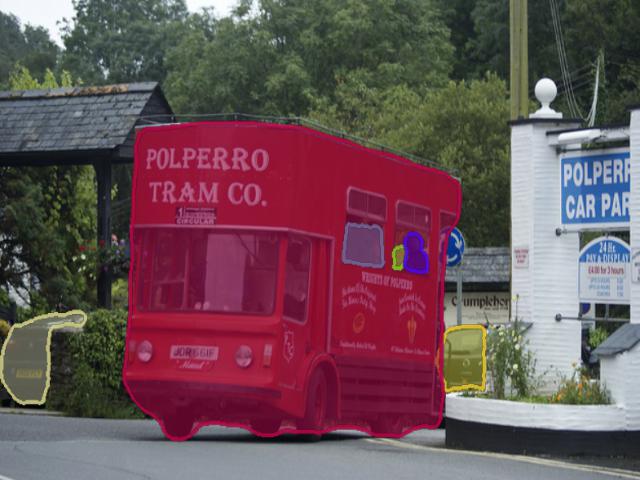}\incs\incg{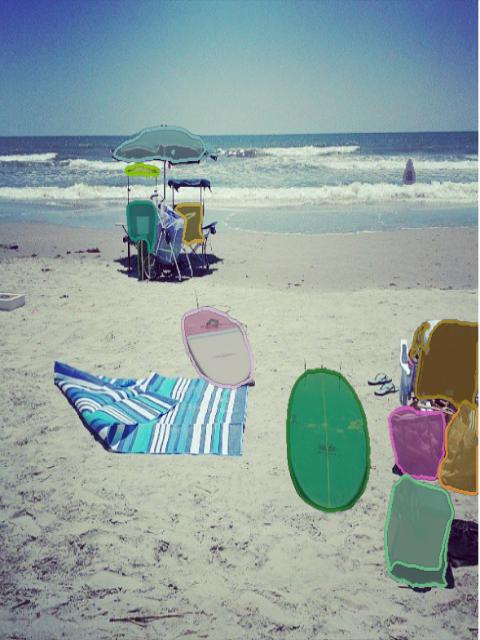}\\
  \incg{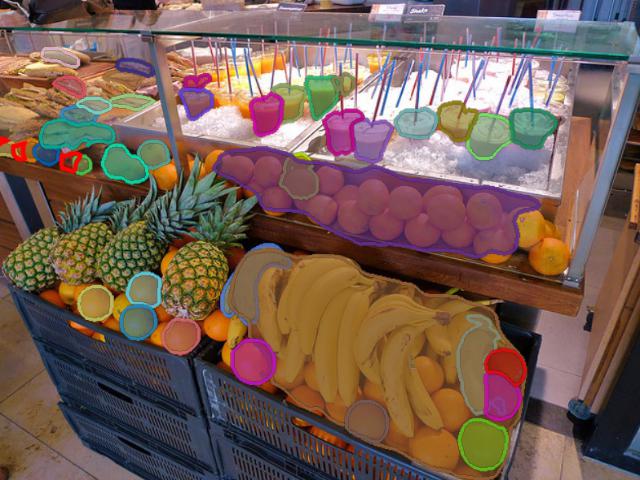}\incs\incg{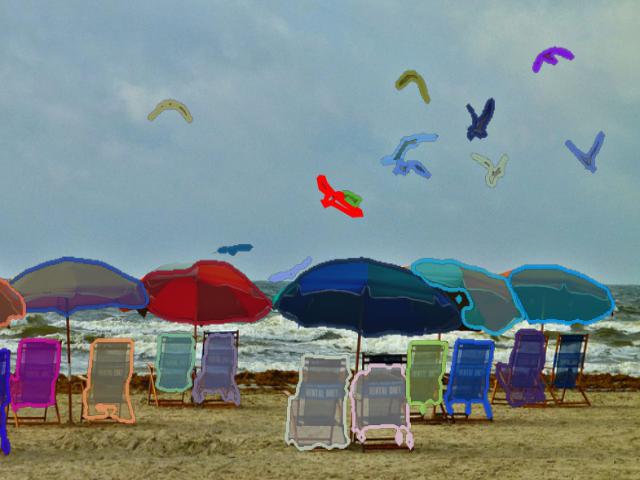}\incs\incg{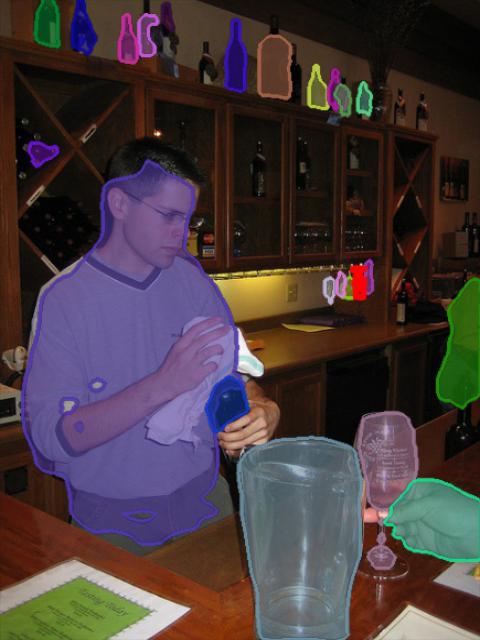}\\
  \incg{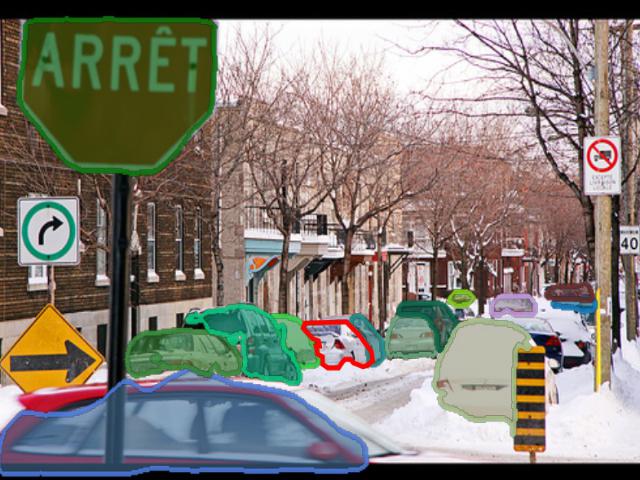}\incs\incg{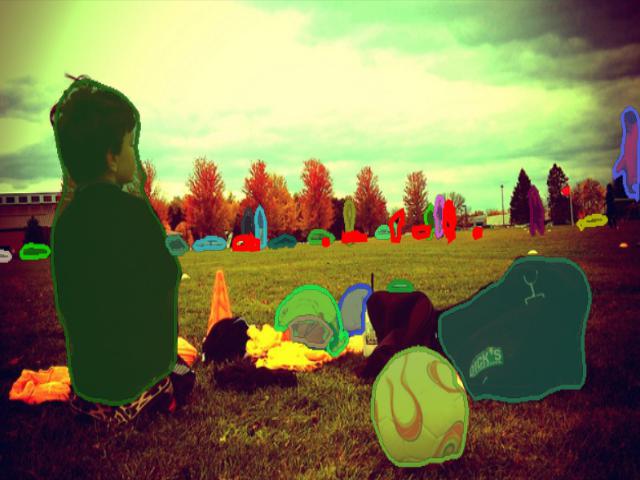}\incs\incg{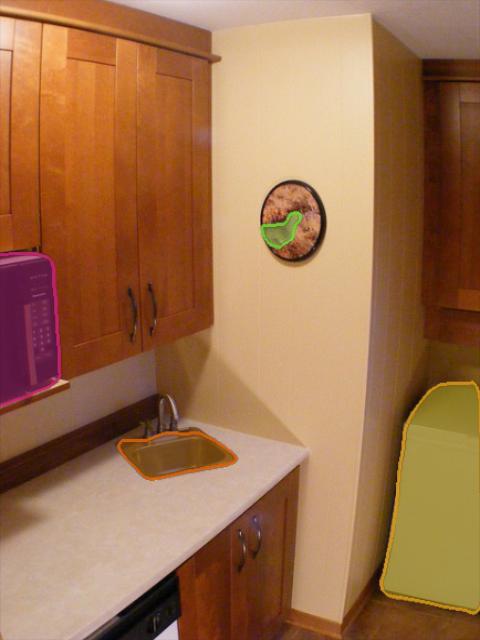}\\
  \incg{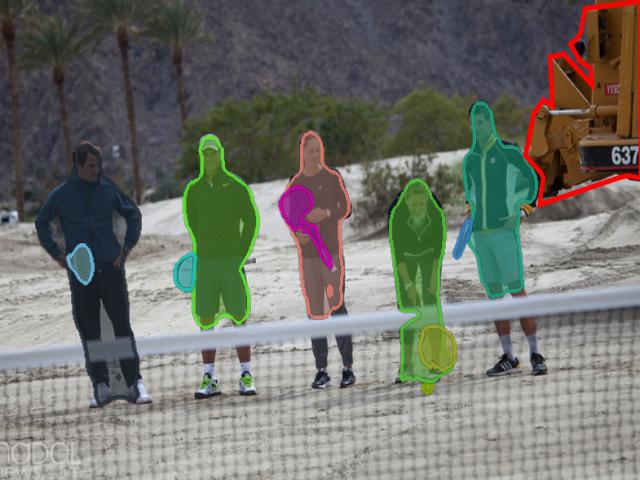}\incs\incg{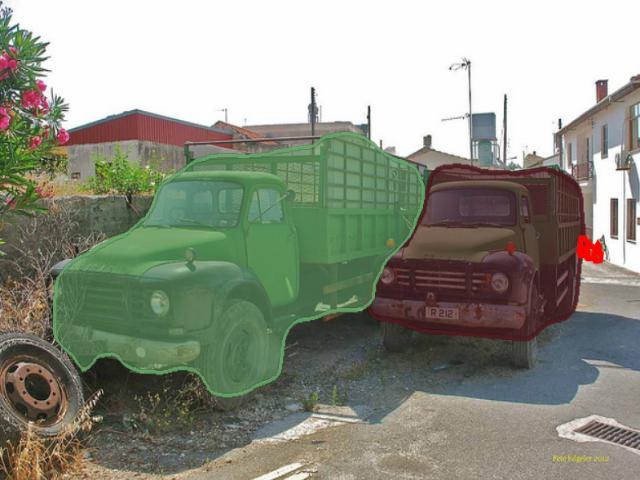}\incs\incg{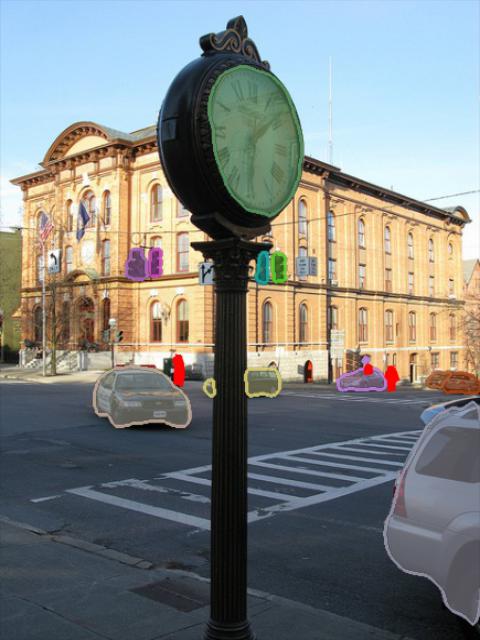}\\
  \incg{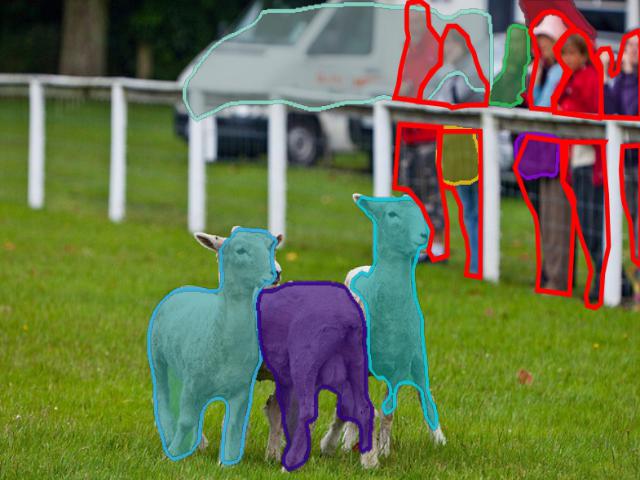}\incs\incg{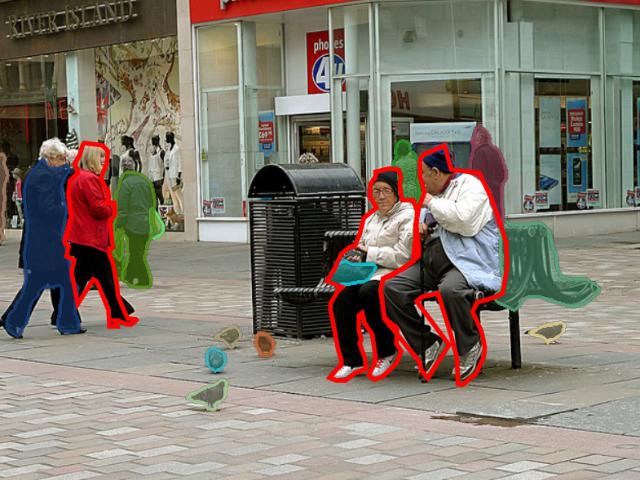}\incs\incg{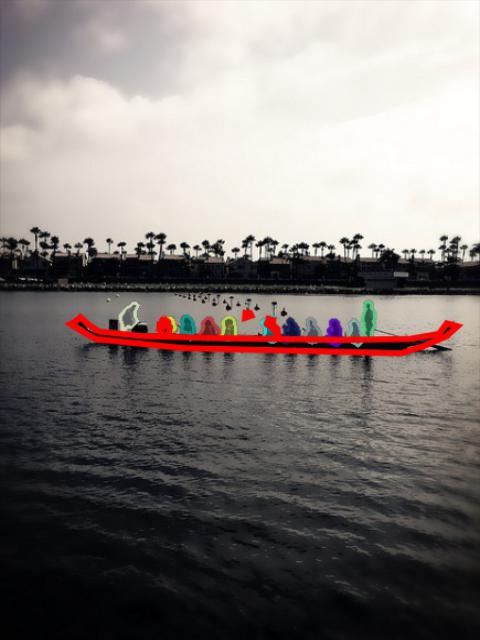}\\
\caption{More selected qualitative results (see also Figure~\ref{fig:qualitative}).}
\label{fig:qualitative:more}
\end{figure}
\clearpage
%##################################################################################################

%##################################################################################################
\begin{figure}[t]
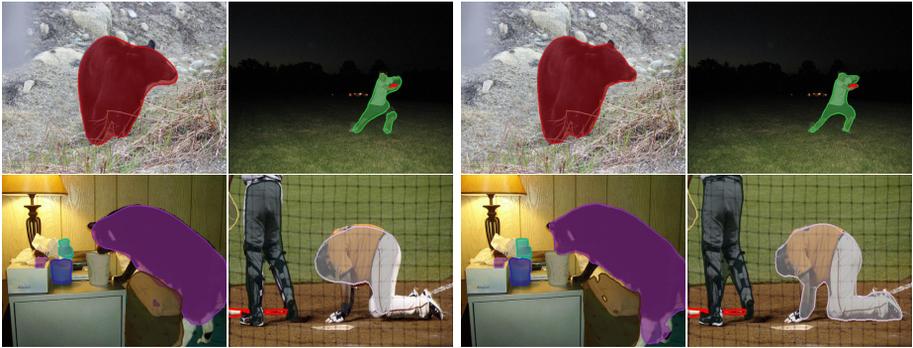
\centering
  \begin{subfigure}[b]{0.495\textwidth}\centering
    \incp{DeepMask}{502}\incs\incp{DeepMask}{589}\\
    \incp{DeepMask}{675}\incs\incp{DeepMask}{999}
    \caption{DeepMask Output}\label{fig:outDeepMask:more}
  \end{subfigure}
  \begin{subfigure}[b]{0.495\textwidth}\centering
    \incp{SharpMask}{502}\incs\incp{SharpMask}{589}\\
    \incp{SharpMask}{675}\incs\incp{SharpMask}{999}
    \caption{SharpMask Output}\label{fig:outSharpMask:more}
  \end{subfigure}%
\caption{More selected qualitative comparisons (see also Figure~\ref{fig:comparison}).}
\label{fig:comparison:more}
\end{figure}
%##################################################################################################

%%%%%%%%%%%%%%%%%%%%%%%%%%%%%%%%%%%%%%%%%%%%%%%%%%%%%%%%%%%%%%%%%%%%%%%%%%%%%%%%%%%%%%%%%%%%%%%%%%%
\bibliographystyle{splncs}\bibliography{bibliography}

\end{document}